\documentclass[final,5p,twocolumn,nopreprintline]{elsarticle}

\usepackage{lineno,hyperref}
\modulolinenumbers[5]

\journal{Pattern Recognition}

\usepackage{amsmath}
\usepackage{amsthm}
\usepackage{amssymb}

\usepackage{afterpage}% http://ctan.org/pkg/afterpage
\newlength{\oldtextfloatsep}

\DeclareMathOperator*{\argmin}{arg\,min}
\DeclareMathOperator*{\argmax}{arg\,max}

\usepackage[ruled,linesnumbered]{algorithm2e}
\usepackage{subfig}

\usepackage[ruled,linesnumbered]{algorithm2e}
\usepackage{subfig}

\usepackage{tikz}
\usetikzlibrary{fit,positioning,arrows,calc,backgrounds,patterns}

\newcommand*\rot{\rotatebox{90}}
\usepackage{booktabs}
\begin{document}

\begin{frontmatter}

\title{Integrated Inference and Learning of Neural Factors\\in Structural Support Vector Machines}

\author[]{Rein Houthooft\corref{mycorrespondingauthor}}
\cortext[mycorrespondingauthor]{Corresponding author}
\ead{rein.houthooft@ugent.be}

\author[]{Filip De Turck}
\ead{filip.deturck@ugent.be}

\address{Ghent University - iMinds, Department of Information Technology\\ Technologiepark 15, Ghent, B-9052, Belgium}

\begin{abstract}
Tackling pattern recognition problems in areas such as computer vision, bioinformatics, speech or text recognition is often done best by taking into account task-specific statistical relations between output variables. In structured prediction, this internal structure is used to predict multiple outputs simultaneously, leading to more accurate and coherent predictions. Structural support vector machines (SSVMs) are nonprobabilistic models that optimize a joint input-output function through margin-based learning. Because SSVMs generally disregard the interplay between unary and interaction factors during the training phase, final parameters are suboptimal. Moreover, its factors are often restricted to linear combinations of input features, limiting its generalization power. To improve prediction accuracy, this paper proposes: (i) Joint inference and learning by integration of back-propagation and loss-augmented inference in SSVM subgradient descent; (ii) Extending SSVM factors to neural networks that form highly nonlinear functions of input features. Image segmentation benchmark results demonstrate improvements over conventional SSVM training methods in terms of accuracy, highlighting the feasibility of end-to-end SSVM training with neural factors.
\end{abstract}

\begin{keyword}
structural support vector machine \sep neural factors \sep structured prediction \sep neural networks \sep image segmentation
%\MSC[2010] 00-01\sep  99-00
\end{keyword}

\end{frontmatter}

%\linenumbers

\section{Introduction}
\label{sec:introduction}

In traditional machine learning, the output consists of a single scalar, whereas in structured prediction, the output can be arbitrarily structured. These models have proven useful in tasks where output interactions play an important role. Examples are image segmentation, part-of-speech tagging, and optical character recognition, where taking into account contextual cues and predicting all output variables at once is beneficial. A widely used framework is the conditional random field (CRF), which models the statistical conditional dependencies between input and output variables, as well as between output variables mutually. However, many tasks only require `most-likely' predictions, which led to the rise of nonprobabilistic approaches. Rather than optimizing the Bayes' risk, these models minimize a structured loss, allowing the optimization of performance indicators directly \cite{Nowozin:2011:SLP:2185833.2185834}. One such model is the structural support vector machine (SSVM) \cite{tsochantaridis2005large} in which a generalization of the hinge loss to multiclass and multilabel prediction is used.

A downside to traditional SSVM training is the bifurcated training approach in which \textit{unary factors} (dependencies of outputs on inputs), and \textit{interaction factors} (mutual output dependencies) are trained sequentially. A unary classification model is optimized, while the interactions are trained post-hoc. However, this two-phase approach is suboptimal, because the errors made during the training of the interaction factors cannot be accounted for during training of the unary classifier. Another limitation is that SSVM factors are linear feature combinations, restricting the SSVM's generalization power. We propose to extend these linearities to highly nonlinear functions by means of multilayer neural networks, to which we refer as \textit{neural factors}. Towards this goal, subgradient descent is extended by combining loss-augmented inference with back-propagation of the SSVM objective error into both unary and interaction neural factors. This leads to better generalization and more synergy between both SSVM factor types, resulting in more accurate and coherent predictions.

Our model is empirically validated by means of the complex structured prediction task of image segmentation on the MSRC-21, KITTI, and SIFT Flow benchmarks. The results demonstrate that integrated inference and learning, and/or using neural factors, improves prediction accuracy over conventional SSVM training methods, such as $N$-slack cutting plane and subgradient descent optimization \cite{Nowozin:2011:SLP:2185833.2185834}. Furthermore, we demonstrate that our model is able to perform on par with current state-of-the-art segmentation models on the MSRC-21 benchmark.

\section{Related work}
\label{sec:related_work}

% Hier wat algemeen werk dat nonlinear factors wilt combineren met structured prediction, maar two-phase en/of linear-chain en/of enkel unary.

% dhungel2014deep doet ook deep learning voor unary potentials, maar blijft werken met gewoon SSVM framework voor pairwise.

Although the combination of neural networks and structured or probabilistic graphical models dates back to the early '90s \cite{bottou1997global, NIPS1989_195}, interest in this topic is resurging. Several recent works introduce nonlinear unary factors/potentials into structured models. For the task of image segmentation, Chen et al.\ \cite{Chen2015} train a convolutional neural network as a unary classifier, followed by the training of a dense random field over the input pixels. Similarly, Farabet et al.\ \cite{farabet-pami-13} combine the output maps of a convolutional network with a CRF for image segmentation, while Li and Zemel \cite{li2014high} propose semisupervised maxmargin learning with nonlinear unary potentials. Contrary to these works, we trade the bifurcated training approach for integrated inference and training of unary and interactions factors. Several works \cite{collobert2011natural, morris2008conditional, Prabhavalkar2010, Yu2009} focus on linear-chain graphs, using an independently trained deep learning model whose output serves as unary input features. Contrary to these works, we focus on more general graphs. Other works suggest kernels towards nonlinear SSVMs \cite{lucchi, bertelli2011kernelized}; we approach nonlinearity by representing SSVM factors by arbitrarily deep neural networks.

% Hier wat werk over gelijkaardig materiaal dat ook tracht backpropagation te gebruiken voor joint inference en learning, maar niet voor pairwise interactions.

Do and Arti\`eres \cite{do2010neural} propose a CRF in which potentials are represented by multilayer networks. The performance of their linear-chain probabilistic model is demonstrated by optical character and speech recognition using two-hidden-layer neural network outputs as unary potentials. Furthermore, joint inference and learning in linear-chain models is also proposed by Peng et al.\ \cite{peng2009conditional}, however, the application to more general graphs remains an open problem \cite{amullerthesis}. Contrary to these works, we popose a nonprobabilistic approach for general graphs by also modeling nonlinear interaction factors. More recently, Schwing and Urtasun \cite{schwing2015fully} train a convolutional network as a unary classifier jointly with a fully-connected CRF for the task of image segmentation, similar to \cite{Tompson2014, krahenbuhl2013parameter}. Chen et al.~\cite{Chen2014} advocate a joint learning and reasoning approach, in which a structured model is probabilistically trained using loopy belief propagation for the task of optical character recognition and image tagging. Other related work includes Domke \cite{domke2013structured} who uses relaxations for combined message-passing and learning.

Other related work aiming to improve conventional SSVMs are the works of Wang et al.\ \cite{wang2013incorporating} and Lin et al.\ \cite{lin2015discriminatively}, in which a hierarchical part-based model is proposed for multiclass object recognition and shape detection, focusing on model reconfigurability through compositional alternatives in And-Or graphs. Liang et al.\ \cite{liang2015deep} propose the use of convolutional neural networks to model an end-to-end relation between input images and structured outputs in active template regression. Xu et al.\ \cite{xu2014compositional} propose the learning of a structured model with multilayer deformable parts for action understanding, while Lu et al.\ \cite{lu2015human} propose a hierarchical structured model for action segmentation.

Many of these works use probabilistic models that maximize the negative log-likelihood, such as \cite{do2010neural, peng2009conditional}. In contrast, this paper takes a nonprobabilistic approach, wherein an SSVM is optimized via subgradient descent. The algorithm is altered to back-propagate SSVM loss errors, based on the ground truth and a loss-augmented prediction into the factors. Moreover, all factors are nonlinear functions, allowing the learning of complex patterns that originate from interaction features.

\section{Methodology}

In this section, essential SSVM background is introduced, after which integrated inference and back-propagation is explained for nonlinear unary factors. Finally, this notion is generalized into an SSVM model using only neural factors which are optimized by an alteration of subgradient descent.

\subsection{Background}
\label{sec:bg}

Traditional classification models are based on a prediction function $f : \mathcal{X} \rightarrow \mathbb{R}$ that outputs a scalar. In contrast, structured prediction models define a prediction function $f : \mathcal{X} \rightarrow \mathcal{Y}$, whose output can be arbitrarily structured. In this paper, this structure is represented by a vector in $\mathcal{Y} = \mathcal{L}^n$, with $\mathcal{L} \subset \mathbb{N}$ a set of class labels. Structured models employ a compatibility function $g : \mathcal{X} \times \mathcal{Y} \rightarrow \mathbb{R}$, parametrized by $w \in \mathbb{R}^D$. Prediction is done by solving the following maximization problem:
\begin{equation}
f(x) = \argmax_{y \in \mathcal{Y}} g(x,y;w).
\label{eq:NPhard}
\end{equation}
This is called inference, i.e., obtaining the most-likely assignment of labels, which is similar to maximum-a-posteriori (MAP) inference in probabilistic models. Because of the combinatorial complexity of the output space $\mathcal{Y}$, the maximization problem in Eq.~\eqref{eq:NPhard} is NP-hard \cite{Chen2014}. Hence, it is important to impose on $g$ some kind of regularity that can be exploited for inference. This can be done by ensuring that $g$ corresponds to a nonprobabilistic factor graph, for which efficient inference techniques exist \cite{Nowozin:2011:SLP:2185833.2185834}. In general, $g$ is linearly parametrized as a product of a weight vector $w$ and a joint feature function $\varphi : \mathcal{X} \times \mathcal{Y} \rightarrow \mathbb{R}^D$.

Commonly, $g$ decomposes as a sum of unary and interaction factors\footnote{Maximizing $g$ corresponds to minimizing the state of a nonprobabilistic factor graph, which factorizes into a product of factors. However, by operating in the log-domain, the state decomposes as a sum of factors.}, in which $\varphi =  [(\varphi_U)^\top, (\varphi_I)^\top]^\top$. The functions $\varphi_U$ and $\varphi_I$ are then sums over all individual joint input-output features of the nodes $\psi_i(y, x)$ and interactions	 $\psi_{ij}(y,x)$ of the corresponding factor graph \cite{Nowozin:2011:SLP:2185833.2185834, lucchi}. For example in the use case of Section~\ref{sec:experiments}, nodes are image regions, while interactions are connections between regions, each with their own joint feature vector. Data samples $(x,y)$ are conform this graphical structure, i.e., $x$ is composed of unary features $x^U$ and interaction features $x^I$. Moreover, the unary and interaction parameters are generally concatenated as $w = [(w_U)^\top, (w_I)^\top]^\top$.

In this formulation, the unary features are defined as
\begin{equation}
\psi_i(y_i,x_i) = \left(\epsilon_i(x)^\top [ y_i = m]\right)^\top_{(m \in \mathcal{L})},
\end{equation}
while the interaction features for 2nd-order (edges) interactions are defined as 
\begin{equation}
\psi_{ij}(y_i, y_j) = \left(\xi_{ij}(x) [ y_i = m \land y_j = n]  \right)^\top_{\left((m,n) \in \mathcal{L}^2\right)},
\label{eq:I}
\end{equation}
with $\epsilon_i(x)$ the unary features corresponding to node $i$ and $\xi_{ij}(x)$ the interaction features corresponding to interaction (edge) $(i,j)$. Similarly, higher-order interaction features can be incorporated by extending this matrix into higher-order combinations of nodes, according to the interactions. In the experiments of this paper, unary features are bag-of-words features corresponding to each superpixel. Interaction features are also bag-of-words, but this time corresponding to all connected superpixels.

In an SSVM the compatibility function is linearly parametrized as $g(x,y;w) = \langle w, \varphi(x,y) \rangle$ and optimized effectively by minimizing an empirical estimate of the regularized structured risk
\begin{equation}
 R(w) + \frac{\lambda}{N} \sum_{n=1}^N \Delta \left( y^n, f(x^n) \right),
\label{eq:srisk}
\end{equation}
with $\Delta : \mathcal{Y} \times \mathcal{Y} \rightarrow \mathbb{R}^+$ a structured loss function for which holds $\forall y, y' \in \mathcal{Y} : \Delta(y,y') \geq 0$, $\Delta(y,y) = 0$, and $\Delta(y',y) = \Delta(y,y')$; $R$ a regularization function; $\lambda$ the inverse of the regularization strength; for a set of $N$ training samples $\{(x^n, y^n)\}_{n \in \{1,\ldots,N\}} \subset \mathcal{X} \times \mathcal{Y}$ that can be decomposed into $V_n$ nodes and $E_n$ interactions. In this paper, we make use of $L_2$-regularization, hence $R(w) = \frac{1}{2}\lVert w \rVert ^2$. Furthermore, in line with our image segmentation use case in Section~\ref{sec:experiments}, the loss function is the class-weighted Hamming distance between two label assignments, or
\begin{equation}
\Delta(y^n, y) = \sum_{i = 1}^{V_n} \eta(y^n_i) [y^n_i \neq y_i],
\label{eq:loss}
\end{equation}
with $[\cdot]$ the Iverson brackets and $V_n$ the number of nodes (i.e., inputs to the unary factors, which corresponds to the number of nodes in the underlying factor graph) in the $n$-th training sample. Contrary to maximum likelihood approaches \cite{do2010neural, Chen2014, krahenbuhl2013parameter}, the Hamming distance allows us to directly maximize performance metrics regarding accuracy. By setting $\eta(y^n_i) = 1$ we can focus on node-wise accuracy, while setting $\eta(y^n_i) = (\sum_{i,n} [y^n_i = y_i])^{-1}$ allows us to focus on class-mean accuracy.

Due to the piecewise nature of the loss function $\Delta$, traditional gradient-based optimization techniques are ineffective for solving Eq.~\eqref{eq:srisk}. However, according to Zhang \cite{zhang2004statistical}, the equations
\begin{equation}
L(w) = \frac{1}{2}\lVert w \rVert ^2 + \frac{\lambda}{N} \sum^N_{n=1} \max\{ \ell(x^n, y^n; w), 0\}, \text{ with}\label{eq:ssvm}
\end{equation}
\begin{multline}
\ell(x^n, y^n; w) =\\ \max_{y \in \mathcal{Y}} [ \Delta(y^n,y) - g(x^n, y^n; w) + g(x^n, y; w) ],
\label{eq:ssvm2}
\end{multline}
define a continuous and convex upper bound for the actual structured risk in Eq.~\eqref{eq:srisk} that can be minimized effectively by solving $\argmin_{w \in \mathbb{R}^D} L(w)$ through numerical optimization \cite{Nowozin:2011:SLP:2185833.2185834, zhang2004statistical}.

\subsection{Integrated back-propagation and inference}

\IncMargin{1em}
\begin{algorithm*}[!t]
%\SetAlgoVlined
\DontPrintSemicolon % Some LaTeX compilers require you to use \dontprintsemicolon instead
\Indentp{-1em}
\KwIn{\# iterations $T$; learning rate curve $\mu / (t_0 + t)$; inverse regularization strength $\lambda$; training samples $\{(x^n,y^n)\}_{n \in \{1,\ldots,N\}}$}
\KwOut{optimized parameters $\theta \in \mathbb{R}^K$ and $w \in \mathbb{R}^{L}$}
\Indentp{1em}
Initialize $w$ to $\vec{0}$ and $\theta$ according to \cite{glorot2010understanding}; the output layer weights are initialized to 0.\;
\For{$1 \leq t \leq T$}{
\For{$1 \leq n \leq N$}{
$z^n \leftarrow \argmax_{y \in \mathcal{Y}} [ \Delta(y^n, y) + \langle w, \varphi_I(x^n, y) \rangle+ f(x^n, y; \theta) ] $\tcp*{\small loss-augmented prediction in Eq.~\eqref{eq:lap}}
\vspace{4pt}
    \eIf(\texttt{\small\hspace{92pt}// $\max$-operation in Eq.~\eqref{eq:ssvm}}){$ \Delta(y^n,y) - g(x^n, y^n; \theta, w) + g(x^n, z^n; \theta, w) > 0$}{
\vspace{4pt}
$\dfrac{ \partial L_n}{\partial w}(\theta, w) \leftarrow w  + \lambda \big( \varphi_I\left(x^n, z^n\right) - \varphi_I(x^n, y^n) \big)$\tcp*{\small standard SSVM subgradient computation \cite{Nowozin:2011:SLP:2185833.2185834}}
\vspace{4pt}
$\nabla_\theta L_n(\theta,w) \leftarrow \theta + \lambda \big( \nabla_\theta f(x^n, z^n; \theta) - \nabla_\theta f(x^n, y^n; \theta) \big)$\tcp*{\small gradient computation as in Eq.~\eqref{eq:backprop}}
}{
$\nabla_\theta L_n(\theta, w) \leftarrow \theta$ and $\dfrac{\partial L_n}{\partial w}(\theta, w) \leftarrow w$\;
}

}
$w \leftarrow w - \dfrac{\mu}{t_0 + t} \dfrac{1}{N} \sum_{n=1}^N \dfrac{\partial L_n}{\partial w}(\theta, w)$ \tcp*{\small update linear interaction factors}
\vspace{4pt}
$\theta \leftarrow \texttt{backprop}\left(\dfrac{1}{N} \sum_{n=1}^N \nabla_\theta L_n(\theta, w)\right)$ \hspace{-40pt}\tcp*{\small update neural unary factors via back-propagation}
    }

\caption{Integrated SSVM subgradient descent with neural unary and linear interaction factors}
\label{algo:duplicate}
%\afterpage{\global\setlength{\textfloatsep}{\oldtextfloatsep}}
\end{algorithm*}
\DecMargin{1em}

Traditional SSVM training methods optimize a joint parameter vector of the unary and interaction factors. However, they restrict these parameters to linear combinations of input features, or allow limited nonlinearity through the addition of kernels. The objective function in case of arbitrary nonlinear factors is often hard to optimize, as many numerical optimization methods require a convex objective function formulation. For example, $N$-slack cutting plane training requires the conversion of the $\max$-operation in Eq.~\eqref{eq:ssvm2} to a set of $N |\mathcal{Y}|$ linear constraints for its quadratic programming procedure \cite{joachims2009cutting}; block-coordinate Frank-Wolfe SSVM optimization \cite{ICML2013_lacoste-julien13} assumes linear input dependencies; the structured perceptron similarly assumes linear parametrization \cite{collins2002discriminative}; and dual coordinate descent focuses on solving the dual of the linear $L_2$-loss in SSVMs \cite{chang2013dual}.

Subgradient descent minimization, as described in \cite{Nowozin:2011:SLP:2185833.2185834,book:shor1985}, is a flexible tool for optimizing Eq.~\eqref{eq:ssvm} as it naturally allows error back-propagation. This algorithm alternates between two steps. First,
\begin{equation}
z^n = \argmax_{y \in \mathcal{Y}} [ \Delta(y^n, y) + \langle w, \varphi(x^n, y)  \rangle ]
\label{eq:lap}
\end{equation}
is calculated for all $N$ training samples, which is called the loss-augmented inference or prediction step, derived from Eq.~\eqref{eq:ssvm2}. In this paper, general inference for determining Eq.~\eqref{eq:NPhard} is approximated via the $\alpha$-expansion \cite{boykov2001fast} algorithm, whose effectiveness has been validated through extensive experiments \cite{Peng20131020}. Loss-augmented prediction as in Eq.~\eqref{eq:lap} is incorporated into this procedure by adding the loss term $\eta(y_i^n) [y_i^n \neq y_i]$ to the unary factors.

Second, these $z$-values are used to calculate a subgradient\footnote{$v \in \mathbb{R}^D$ is a subgradient of $f : \mathbb{R}^D \rightarrow \mathbb{R}$ in a point $p_0$ if $f(p) - f(p_0) \geq \langle v, p - p_0 \rangle$. Due to its piecewise continuous nature, Eq.~\eqref{eq:ssvm} is nondifferentiable in some points, hence we are forced to rely on subgradients.} of Eq.~\eqref{eq:ssvm} as
$\frac{1}{N}[w  + \lambda \left( \varphi\left(x^n, z^n\right) - \varphi(x^n, y^n) \right)]$
for each sample $(x^n,y^n)$, in order to update $w$. Traditional SSVMs assume that $g(x, y; w) = \langle w, \varphi(x, y) \rangle$ in which $\varphi$ is a predefined joint input-output feature function. Commonly, this joint function is made up of the outputs of a nonlinear `unary' classifier $C : \mathcal{X} \rightarrow [0,1]^{|\mathcal{L}|}$, such that $\varphi_U(x,y)$ becomes $\varphi_U(C(x),y)$ \cite{Houthooft-aaai-2016}. This classifier is trained upfront, based on the different unary inputs corresponding to each node in the underlying factor graph. Due to the linear definition of $g$, the SSVM model is learning linear combinations of these classifier outputs as its unary factors. In general, the interaction factors are not trained through a separate classifier, and are thus linear combinations of the interaction features directly.

We propose to replace the pretraining of a nonlinear unary classifier, and the transformation of its outputs through linear factors, by the direct optimization of nonlinear unary factors. In particular, the unary part of $g$ is represented by a sum $f$ of outputs of an adapted neural network which models factor values. To achieve this, the loss-augmented prediction step defined in Eq.~\eqref{eq:lap} is altered to
\begin{equation}
z^n = \argmax_{y \in \mathcal{Y}} [ \Delta(y^n, y) + \langle w, \varphi_I(x^n, y) \rangle + f(x^n, y; \theta)],\label{eq:lamap2}
\end{equation}
in which $\varphi_I$ represents the joint interaction feature function as described in Section~\ref{sec:bg} and Eq.~\eqref{eq:I}. Eq.~\eqref{eq:lamap2} is calculated similarly to Eq.~\eqref{eq:lap} through $\alpha$-expansion by encoding the loss term into the unary factors.

The compatibility function thus becomes $g(x, y; \theta, w) = \langle w, \varphi_I(x, y) \rangle + f(x, y; \theta)$. The calculation of $\frac{\partial L}{\partial w}$, originally defined as the subderivative of the objective function in Eq.~\eqref{eq:ssvm}, remains unaltered. However, we can no longer assume that $\frac{\partial L}{\partial \theta}$ conforms to the definition of a subgradient due to its nonconvexity. However, we can calculate
%\begin{multline}
%\frac{\partial L}{\partial \theta}(\theta, w) =\\ \theta + \frac{\lambda}{N} \sum_{n \in \mathcal{N}} \left( \frac{\partial f}{\partial \theta}\left(x^n, z^n; \theta\right) - \frac{\partial f}{\partial \theta}\left(x^n, y^n; \theta \right) \right),\label{eq:backprop}
%\end{multline}
\begin{multline}
\nabla_\theta L (\theta, w) =\\ \theta + \frac{\lambda}{N} \sum_{n \in \mathcal{N}} \left( \nabla_\theta f \left(x^n, z^n; \theta\right) - \nabla_\theta f \left(x^n, y^n; \theta \right) \right),\label{eq:backprop}
\end{multline}
with $\mathcal{N}$ the set of indices corresponding to training samples for which $\ell(x^n,y^n;w) > 0$ in Eq.~\eqref{eq:ssvm2}, for a particular loss-augmented prediction $z^n$. In case $\ell(x^n,y^n;w) = 0$, we set $\nabla_\theta L = \theta$. This gradient incorporates the loss-augmented prediction of Eq.~\eqref{eq:lamap2} and is back-propagated through the underlying network to adjust each element of $\theta$. The altered subgradient descent method is shown in Algorithm~\ref{algo:duplicate}. Herein, $L_n$ represents the objective function for the $n$-th training sample, i.e., $L_n(\theta, w) = \frac{1}{N}[\frac{1}{2} \lVert w \rVert^2 + \lambda( \Delta(y^n,z^n) - g(x^n,y^n;\theta, w) + g(x^n,z^n;\theta, w)) ]$.

\IncMargin{1em}
\begin{algorithm*}[!t]
\DontPrintSemicolon % Some LaTeX compilers require you to use \dontprintsemicolon instead
%\SetAlgoVlined
\Indentp{-1em}
\KwIn{\# iterations $T$; learning rate; inverse regularization strength $\lambda$; training set $\{(x^n,y^n)\}$}
\KwOut{optimized parameters $\theta \in \mathbb{R}^K$ and $\gamma \in \mathbb{R}^M$}
\Indentp{1em}
Initialize $\theta$ and $\gamma$ according to \cite{glorot2010understanding}; the weights of the output layers are initialized to 0.\;
\For{$1 \leq t \leq T$}{
\For{$1 \leq n \leq N$}{
$z^n \leftarrow \argmax_{y \in \mathcal{Y}} [ \Delta(y^n, y) + f(x^n, y; \theta) + h(x^n, y; \gamma) ] $\;
\vspace{4pt}
    \eIf{$ \Delta(y^n,y) - g(x^n, y^n; \theta, \gamma) + g(x^n, z^n; \theta, \gamma) > 0$}{
\vspace{4pt}
$\nabla_\theta L_n (\theta, \gamma) \leftarrow \theta + \lambda \Big( \nabla_\theta f (x^n, z^n; \theta) - \nabla_\theta (x^n, y^n; \theta) \Big)$ \\ \hspace{80pt} and $\ \ \nabla_\gamma L_n (\theta, \gamma) \leftarrow \gamma + \lambda \Big( \nabla_\gamma h (x^n, z^n; \gamma) - \nabla_\gamma h (x^n, y^n; \gamma) \Big)$\;
}{
$\nabla_\theta L_n(\theta, \gamma) \leftarrow \theta$ and $\nabla_\gamma L_n(\theta, \gamma) \leftarrow \theta$\;

}
  }
$\theta \leftarrow \texttt{backprop}\left(\dfrac{1}{N} \sum_{n=1}^N \nabla_\theta L_n (\theta, \gamma)\right)$ and $\gamma \leftarrow \texttt{backprop}\left(\dfrac{1}{N} \sum_{n=1}^N \nabla_\gamma L_n (\theta, \gamma)\right)$\; \label{algo:backprop}

}
\caption{Integrated SSVM subgradient descent with both unary and interaction neural factors}
\label{algo:duplicate2}
\end{algorithm*}
\DecMargin{1em}

In contrast to gradient descent, subgradient methods \cite{Nowozin:2011:SLP:2185833.2185834,book:shor1985} do not guarantee the lowering of the objective function value in each step. Therefore, the current best value $L_*^{(t)} = \min \{ L_*^{(t-1)}, L(w^{(t)}) \}$ is memorized in each iteration $t$, along with the corresponding parameter values $(w_*,\theta_*)$. As such, the objective value $L_*$ decreases at each step as $L_*^{(t)} = \min \{ L(w^{(1)}), \ldots, L(w^{(t)}) \}$. This update rule is omitted from Algorithm~\ref{algo:duplicate} to improve readability.

Because the loss terms in Eq.~\eqref{eq:ssvm2} are no longer affine input transformations due to the introduced nonlinearities of the neural network, we can no longer assume Eq.~\eqref{eq:ssvm} to be convex, as is the case for conventional SSVMs. Although theoretical guarantees can be made for the convergence of (sub)gradient methods for convex functions \cite{nedic2001convergence}, and particular classes of nonconvex functions \cite{bagirov2013subgradient}, no such guarantees can be made for arbitrary nonconvex functions \cite{ngiam2011optimization}. The problem of optimizing highly nonconvex functions is studied extensively in neural network gradient descent literature. However, it has been demonstrated that nonconvex objectives can be minimized effectively due to the high dimensionality of the neural network parameter space \cite{pascanu2014saddle}. Dauphin et al.\ \cite{dauphin2014identifying} show that saddle points are much likelier than local minima in multilayer neural network objective landscapes. In particular, the ratio of saddle points to local minima increases exponentially with the parameter dimensionality. Several methods exists to avoid these these saddle points, e.g., momentum \cite{sutskever2013importance}. Furthermore, Dauphin et al.\ \cite{dauphin2014identifying} show, based on random matrix theory, that the existing local minima are very close to the global minimum of the objective function. This can be understood intuitively as the probability that all directions surrounding a local minimum lead upwards is very small, making local minima not an issue in general. The empirical results presented in Section~\ref{sec:results_disc} reinforce this believe by demonstrating that the regularized objective function can still be minimized effectively, as we achieve accurate predictions.

As described in Algorithm~\ref{algo:duplicate}, the (sub)gradient is defined over whole data samples, which each consist of multiple nodes. $f$ thus models the unary part of the compatibility function $g$, which is a sum of the $V_n$ unary factors. Therefore, the function $f(x,y;\theta)$ decomposes as a sum of \textit{neural unary factors}
\begin{equation}
f(x,y;\theta) = \sum_{i=1}^{V_n} f^\ast(x_{i}^U;\theta)_{\displaystyle y_i},
\label{eq:nn}
\end{equation}
with $x^U$ the unary features in $x$. The nonlinear function  $f^\ast : \mathcal{X} \rightarrow \mathbb{R}^{|\mathcal{L}|}$ is a multiclass multilayer neural network parametrized by $\theta \in \mathbb{R}^K$, whose inputs are features corresponding to the $V_n$ different nodes. It forms a template for the neural unary factors. In this network $f^\ast(x_i^U;\theta)$, the softmax-function is removed from the output layer, such that it matches the unary factor range $\mathbb{R}^{|\mathcal{L}|}$. The argument $y$ of the joint feature function is used as an index $y_i$ to select a particular output unit.

\subsection{Neural interaction factors}
\label{sec:all}

In this section we extend the notion of nonlinear factors beyond the integration of the training of a unary classifier. We now also replace the linear interaction part $\langle w, \varphi_I(x,y)\rangle$ of the compatibility function $g$ with a function $h(x,y;\gamma)$ that decomposes as a sum of \textit{neural interaction factors}
\begin{equation}
h(x,y;\gamma) = \sum_{i=1}^{E_n} h^\ast(x_{i}^I; \gamma)_{\displaystyle \mathcal{N}_i(y)},
\end{equation}
with $x^I$ the interaction features in $x$, $\mathcal{N}_i(y)$ the combination of node labels in the $i$-th interaction, and $E_n$ the number of interactions in the $n$-th training sample. The function $h^\ast: \mathcal{X} \rightarrow \mathbb{R}^{|\mathcal{L}|^Q}$ is parametrized by $\gamma \in \mathbb{R}^M$, and forms a template for the interaction factors. Herein, $Q$ depends on the interaction order, e.g., $Q=2$ in the Section~\ref{sec:experiments} use case as connections between nodes are then edges. Interaction factors are generally not trained upfront. However, neural interaction factors are useful as they can extract complexer interaction patterns, and thus transcend the limited generalization power of linear combinations. In image segmentation for example, interaction features consisting of vertical gradients and a $90^\circ$-angle can indicate that the two connected nodes belong to the same class. The loss-augmented inference step in Eq.~\eqref{eq:lamap2} is now adapted to
\begin{equation}
z^n = \argmax_{y \in \mathcal{Y}}[ \Delta(y^n, y) + f(x^n, y; \theta) + h(x^n, y; \gamma)],
\end{equation}
while the compatibility function becomes $g(x, y; \theta, \gamma) = f(x, y; \theta) + h(x, y; \gamma)$. The two distinct models $f$ and $h$ are trained in a similar fashion to the method described in Algorithm~\ref{algo:duplicate}, as depicted in Algorithm~\ref{algo:duplicate2}. Notice that this method can easily be adjusted for batch or online learning by adapting and moving the weight updates at line~\ref{algo:backprop} into the inner loop.

Like the unary function $f^*$ in Eq.~\eqref{eq:nn}, $h^*(x_i^I;\gamma)$ is a multiclass multilayer neural network in which the top softmax-function is removed, shared among all $E_n$ interaction factors. The output layer dimension matches the number of interaction label combinations, $|\mathcal{L}|^Q$ in the most general case. For example in image segmentation, for a problem with symmetric edge features, the number of output units in $h^\ast$ is $\frac{1}{2}|\mathcal{L}| (|\mathcal{L}|+1)$, which all represent different states for a particular interaction factor (in this case the interactions are undirected edges, thus $\mathcal{N}_i(y)$ consists of the $i$-th edge's incident nodes).

The resulting structured predictor no longer requires two-phase training in which linear interaction factors are combined with the upfront training of a unary classifier, whose output is transformed linearly into unary factor values. It makes use of highly nonlinear functions for all SSVM factors, by way of multilayer neural networks, using an integration of loss-augmented inference and back-propagation in a subgradient descent framework. This allows the factors to generalize strongly while being able to mutually adapt to each other's parameter updates, leading to more accurate predictions.

\section{Experiments}
\label{sec:experiments}

\begin{figure*}[!tp]
\centering
 \captionsetup[subfigure]{labelformat=empty}
\subfloat[]{
   \raisebox{11pt}{\rotatebox[origin=t]{90}{$\color{white}|$GT}}   %%% 0.75in is half of figure height 1.5in
\hspace{2pt}
 \subfloat[]{
   \includegraphics[height=45px]{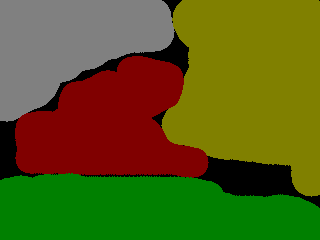}
}\hspace{2pt}
 \subfloat[]{
   \includegraphics[height=45px]{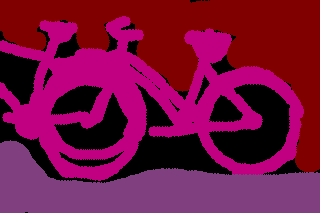}
}\hspace{2pt}
 \subfloat[]{
   \includegraphics[height=45px]{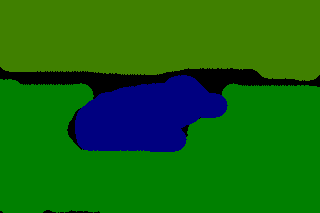}
}\hspace{2pt}
 \subfloat[]{
   \includegraphics[height=45px]{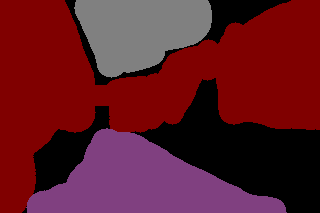}
}\hspace{2pt}
 \subfloat[]{
   \includegraphics[height=45px]{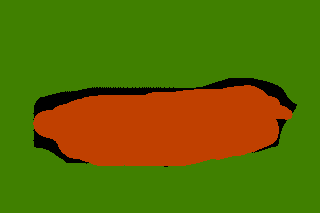}
}\hspace{25pt}
 \subfloat[]{
   \includegraphics[height=45px]{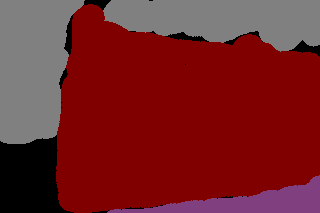}
}}
\vspace{-45pt}
\\
\subfloat[]{
   \raisebox{11pt}{\rotatebox[origin=t]{90}{$\color{white}|$SGD}}   %%% 0.75in is half of figure height 1.5in
\hspace{2pt}
 \subfloat[]{
   \includegraphics[height=45px]{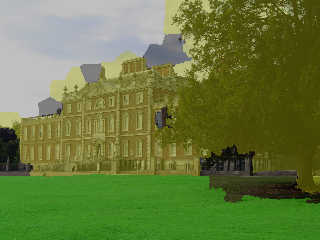}
}\hspace{2pt}
 \subfloat[]{
   \includegraphics[height=45px]{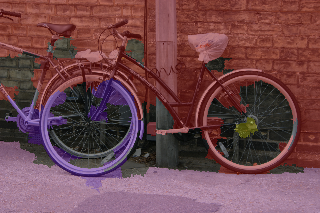}
}\hspace{2pt}
 \subfloat[]{
   \includegraphics[height=45px]{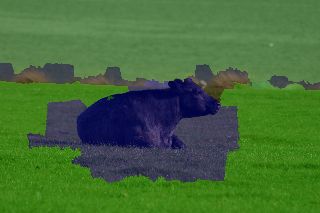}
}\hspace{2pt}
 \subfloat[]{
   \includegraphics[height=45px]{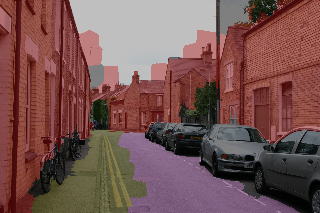}
}\hspace{2pt}
 \subfloat[]{
   \includegraphics[height=45px]{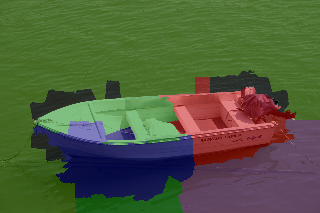}
}\hspace{25pt}
 \subfloat[]{
   \includegraphics[height=45px]{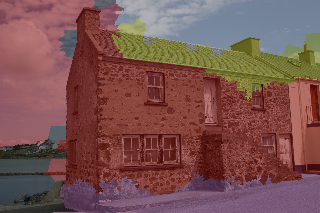}
}}
\vspace{-45pt}
\\
\subfloat[]{
   \raisebox{11pt}{\rotatebox[origin=t]{90}{$\color{white}|$int+nrl}}   %%% 0.75in is half of figure height 1.5in
\hspace{2pt}
 \subfloat[]{
   \includegraphics[height=45px]{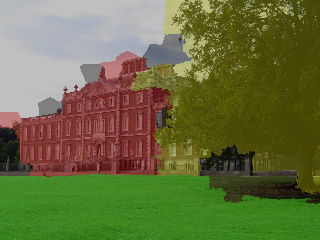}
}\hspace{2pt}
 \subfloat[]{
   \includegraphics[height=45px]{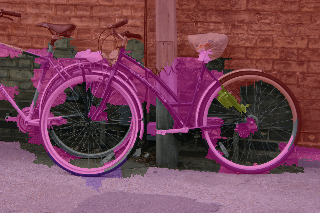}
}\hspace{2pt}
 \subfloat[]{
   \includegraphics[height=45px]{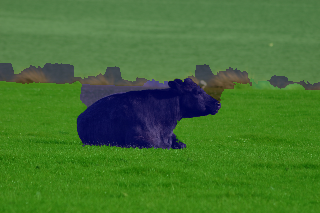}
}\hspace{2pt}
 \subfloat[]{
   \includegraphics[height=45px]{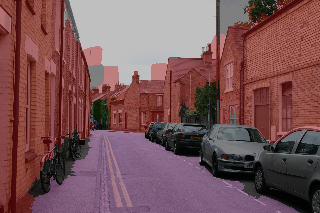}
}\hspace{2pt}
 \subfloat[]{
   \includegraphics[height=45px]{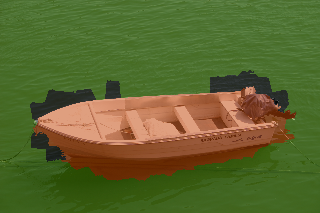}
}\hspace{25pt}
 \subfloat[]{
   \includegraphics[height=45px]{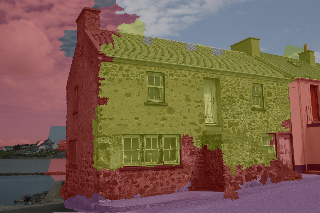}
}}
\vspace{-17pt}
\caption{Illustrative examples of the performance of SGD and int+nrl on several MSRC-21 test images. Integrated training with neural factors improves classification accuracy over subgradient descent. The last column presents a case in which our model fails to outperform SGD.}
\label{fig:results_qualitative}
\end{figure*}

\begin{figure*}[!tp]
\centering
 \captionsetup[subfigure]{labelformat=empty}
\subfloat[]{
   \raisebox{11pt}{\rotatebox[origin=t]{90}{$\color{white}|$GT}}   %%% 0.75in is half of figure height 1.5in
\hspace{2pt}
 \subfloat[]{
   \includegraphics[width=145px]{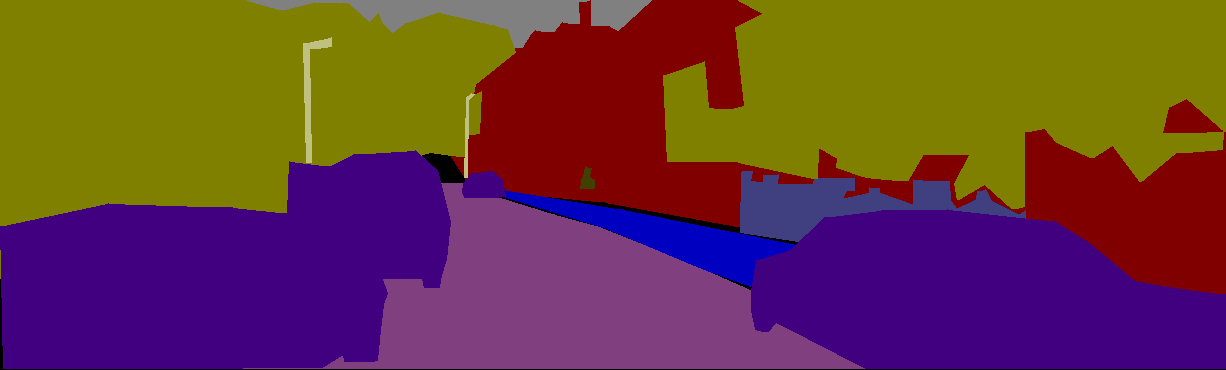}
}\hspace{2pt}
 \subfloat[]{
   \includegraphics[width=145px]{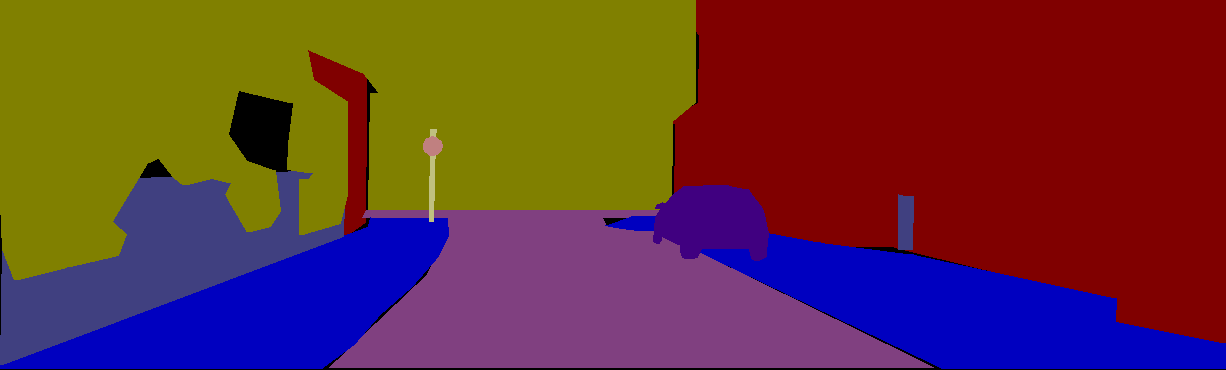}
}\hspace{25pt}
 \subfloat[]{
   \includegraphics[width=145px]{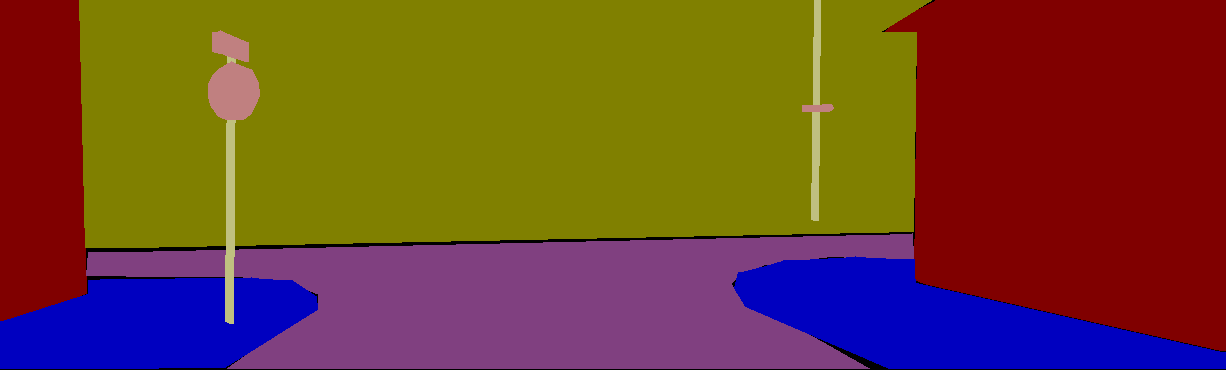}
}}
\vspace{-45pt}
\\
\subfloat[]{
   \raisebox{11pt}{\rotatebox[origin=t]{90}{$\color{white}|$SGD}}   %%% 0.75in is half of figure height 1.5in
\hspace{2pt}
 \subfloat[]{
   \includegraphics[width=145px]{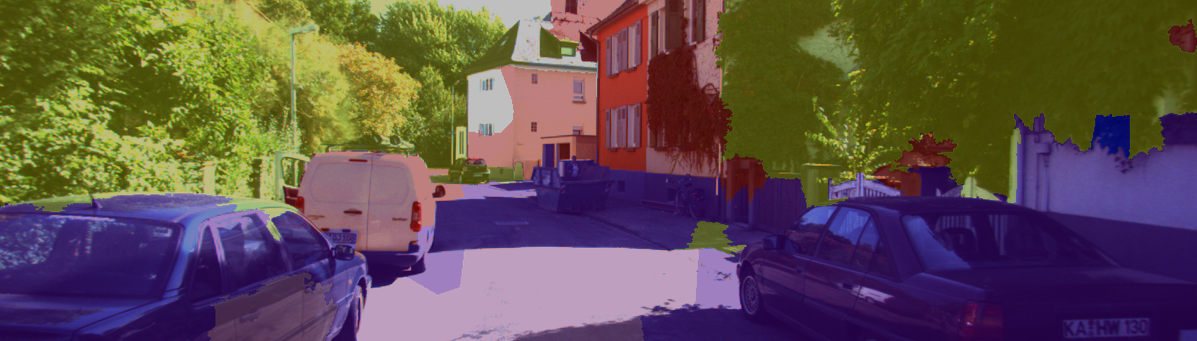}
}\hspace{2pt}
 \subfloat[]{
   \includegraphics[width=145px]{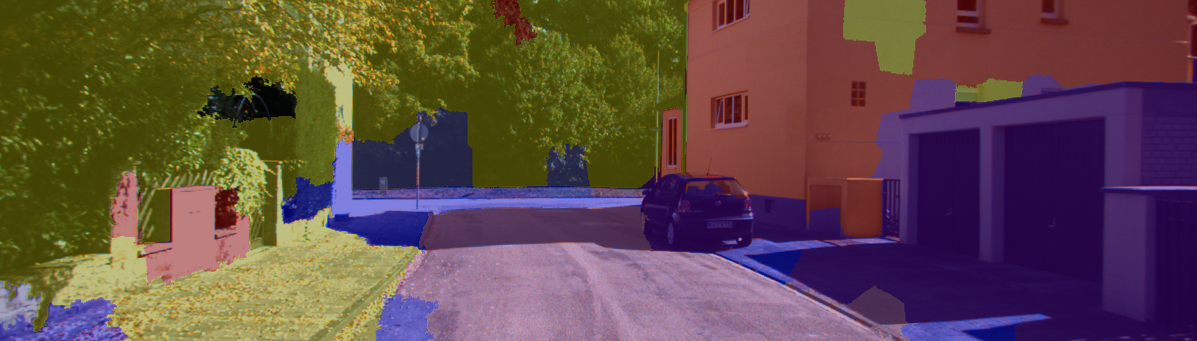}
}\hspace{25pt}
 \subfloat[]{
   \includegraphics[width=145px]{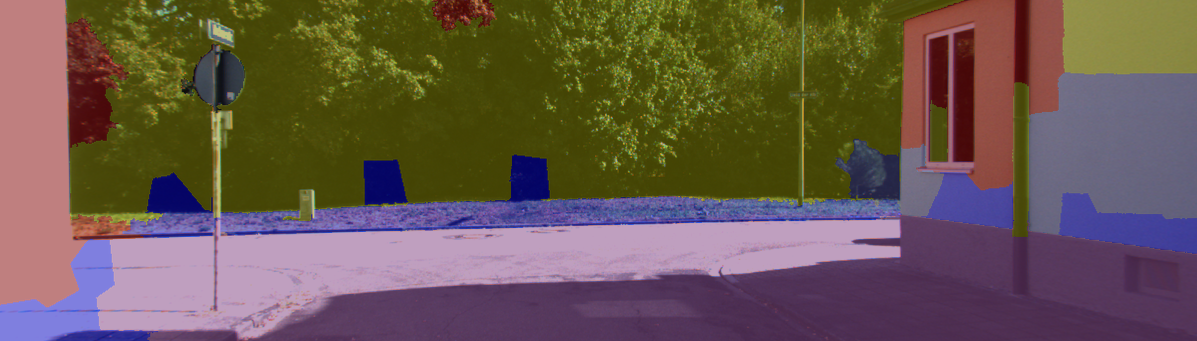}
}}
\vspace{-45pt}
\\
\subfloat[]{
   \raisebox{11pt}{\rotatebox[origin=t]{90}{$\color{white}|$int+nrl}}   %%% 0.75in is half of figure height 1.5in
\hspace{2pt}
 \subfloat[]{
   \includegraphics[width=145px]{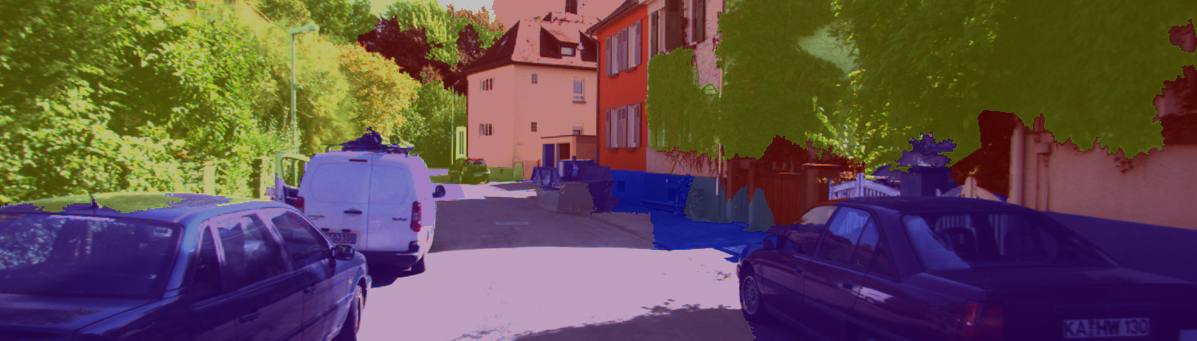}
}\hspace{2pt}
 \subfloat[]{
   \includegraphics[width=145px]{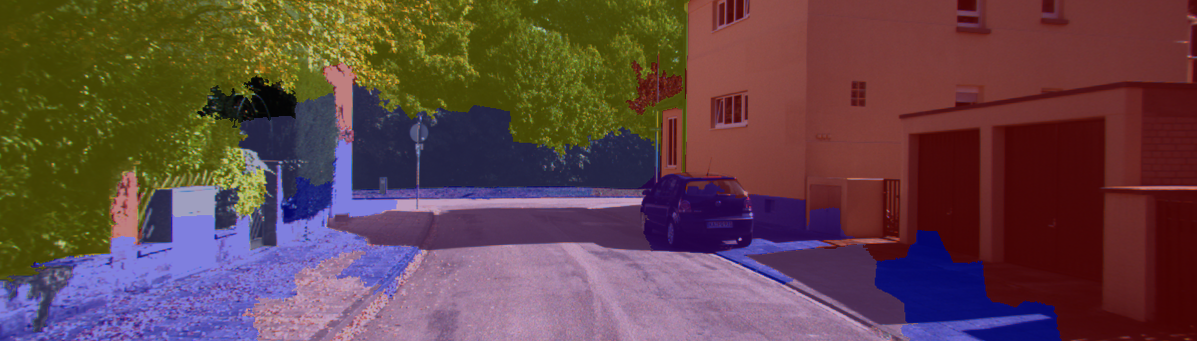}
}\hspace{25pt}
 \subfloat[]{
   \includegraphics[width=145px]{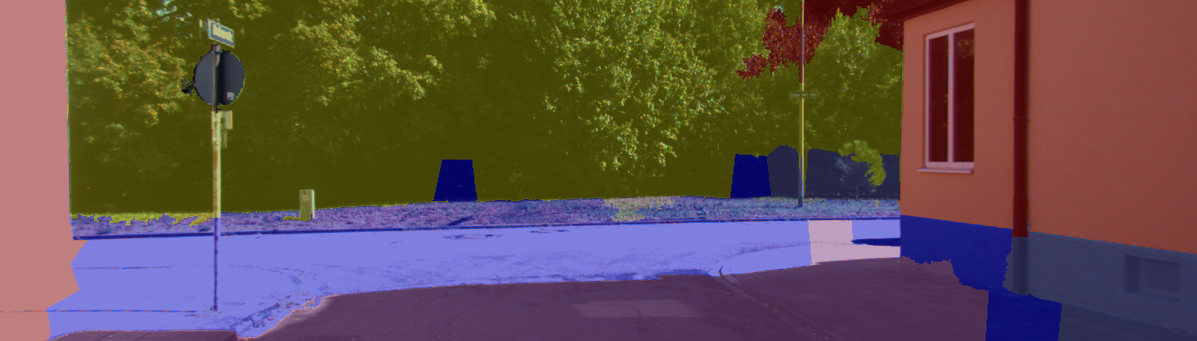}
}}
\vspace{-17pt}
\caption{Illustrative examples of the performance of SGD and int+nrl on several KITTI test images. Integrated training with neural factors improves classification accuracy over subgradient descent. The last column presents a case in which our model fails to outperform SGD.}
\label{fig:kitti_results}
\end{figure*}

\begin{figure*}[!tp]
\centering
 \captionsetup[subfigure]{labelformat=empty}
\subfloat[]{
   \raisebox{17pt}{\rotatebox[origin=t]{90}{$\color{white}|$GT}}   %%% 0.75in is half of figure height 1.5in
\hspace{2pt}
 \subfloat[]{
   \includegraphics[height=55px]{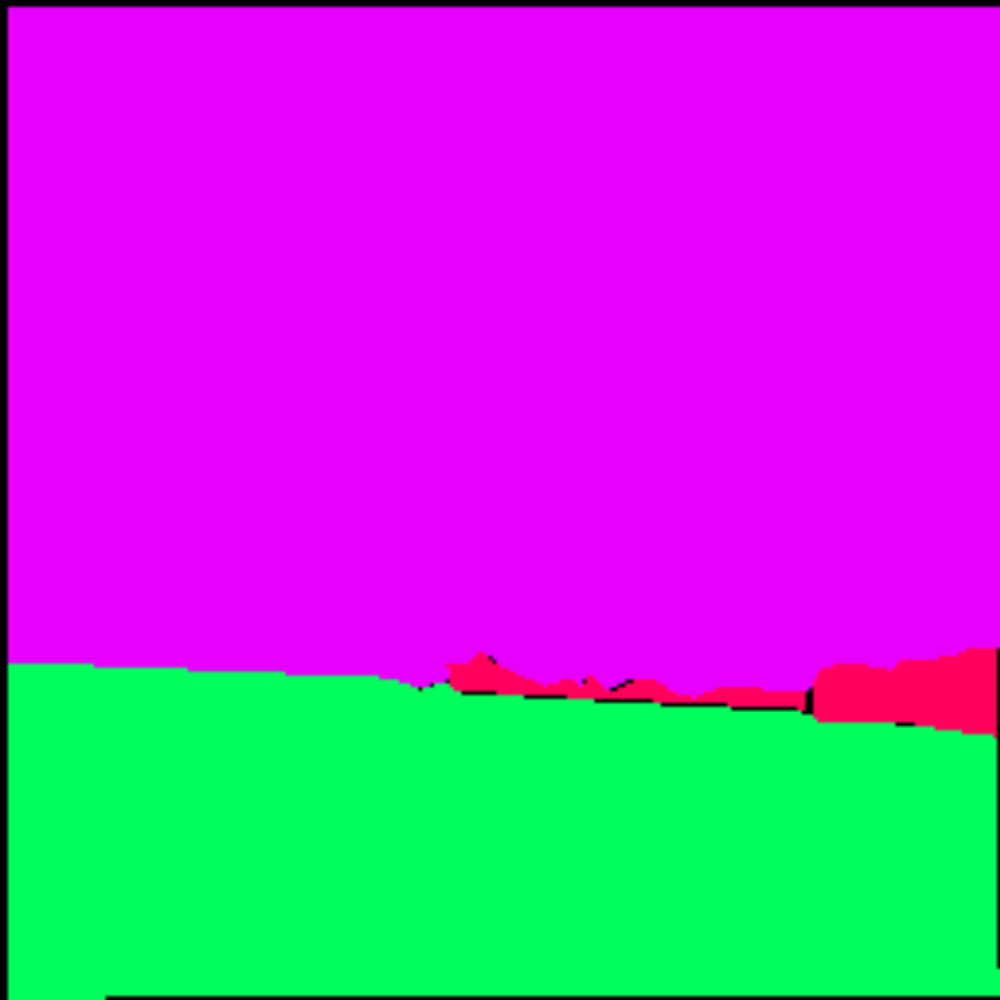}
}\hspace{2pt}
 \subfloat[]{
   \includegraphics[height=55px]{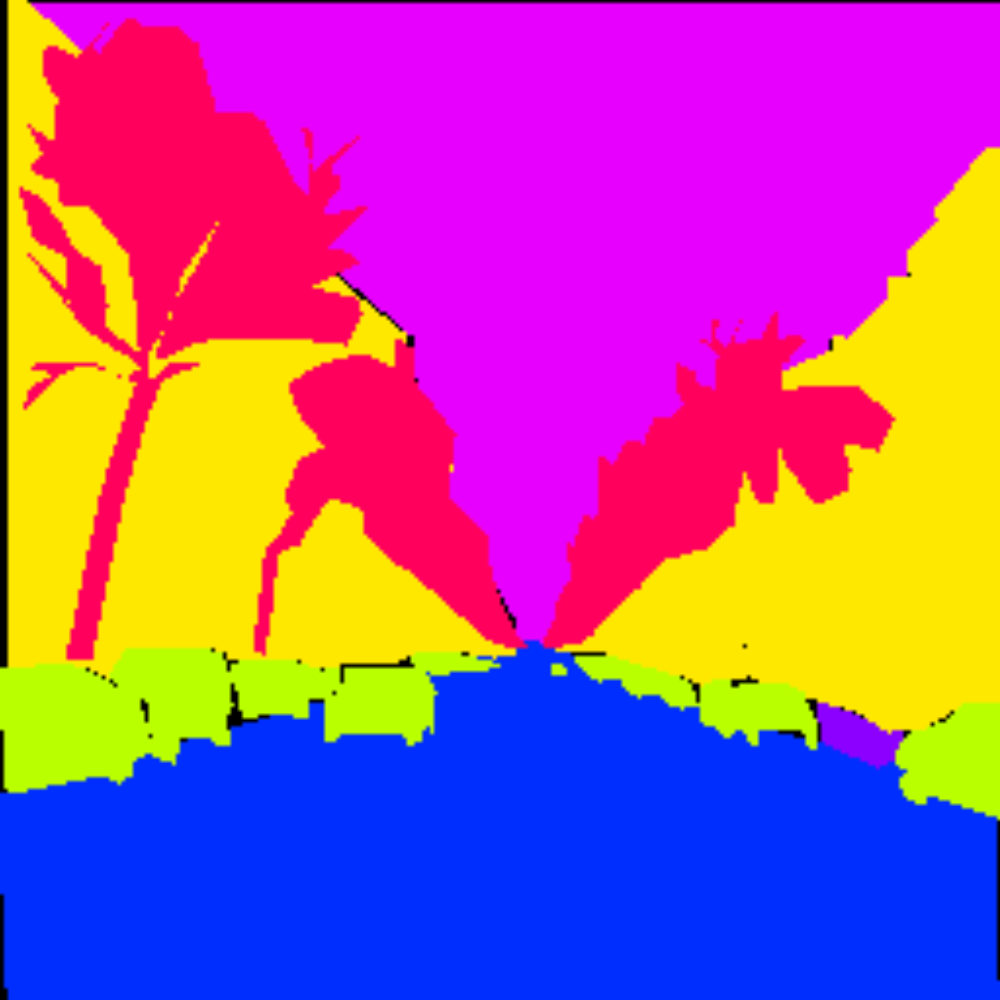}
}\hspace{2pt}
 \subfloat[]{
   \includegraphics[height=55px]{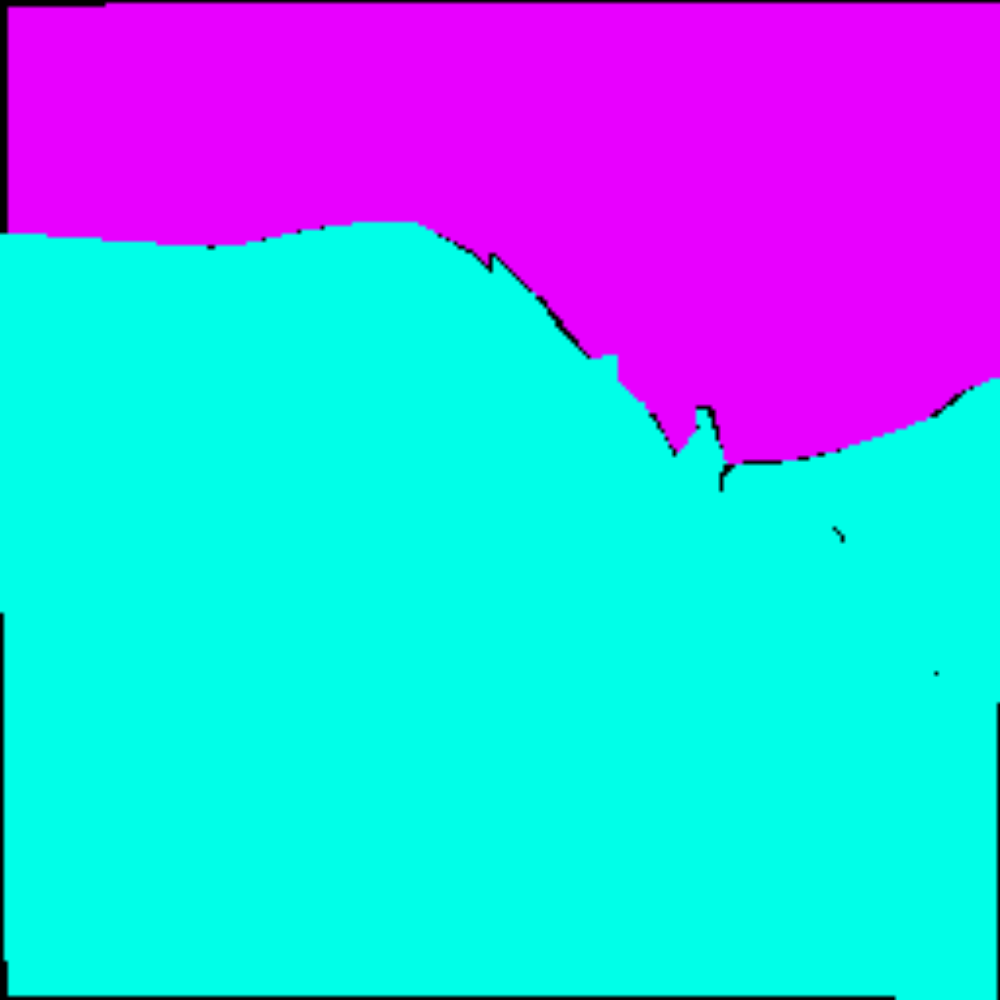}
}\hspace{2pt}
 \subfloat[]{
   \includegraphics[height=55px]{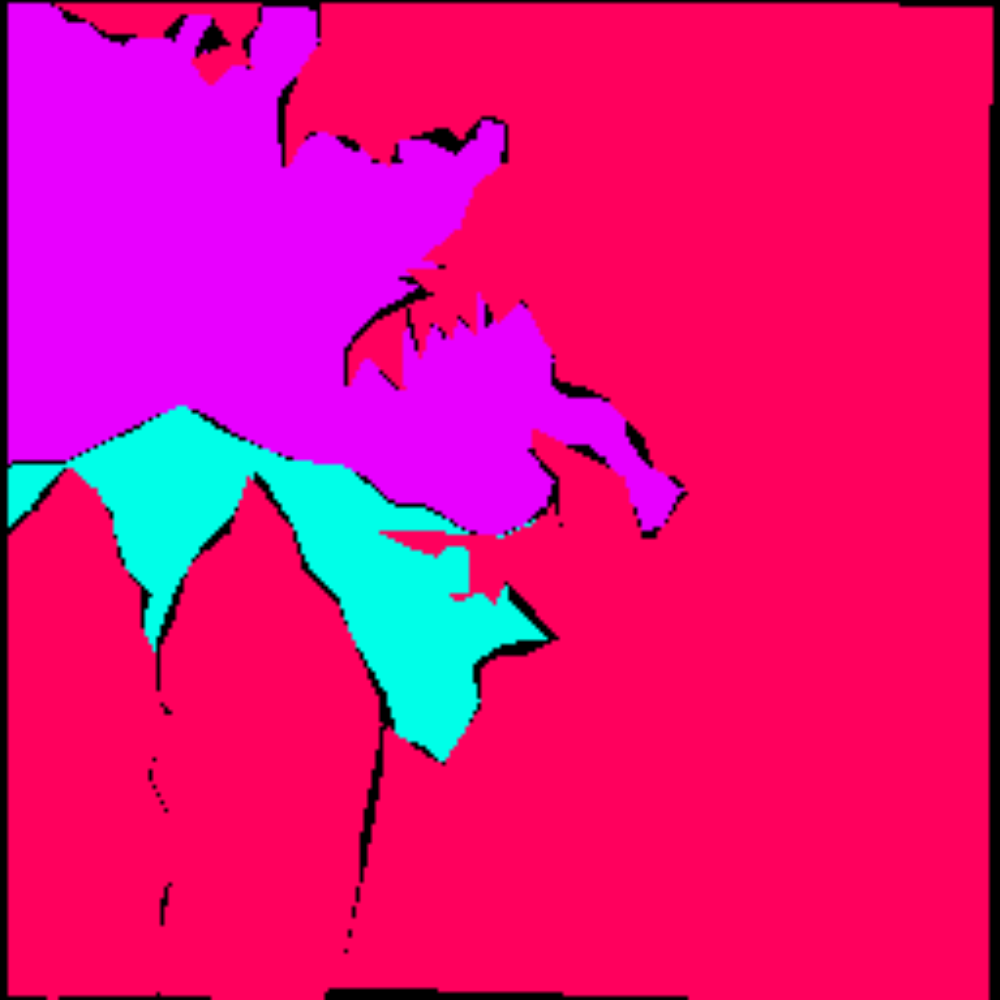}
}\hspace{2pt}
 \subfloat[]{
   \includegraphics[height=55px]{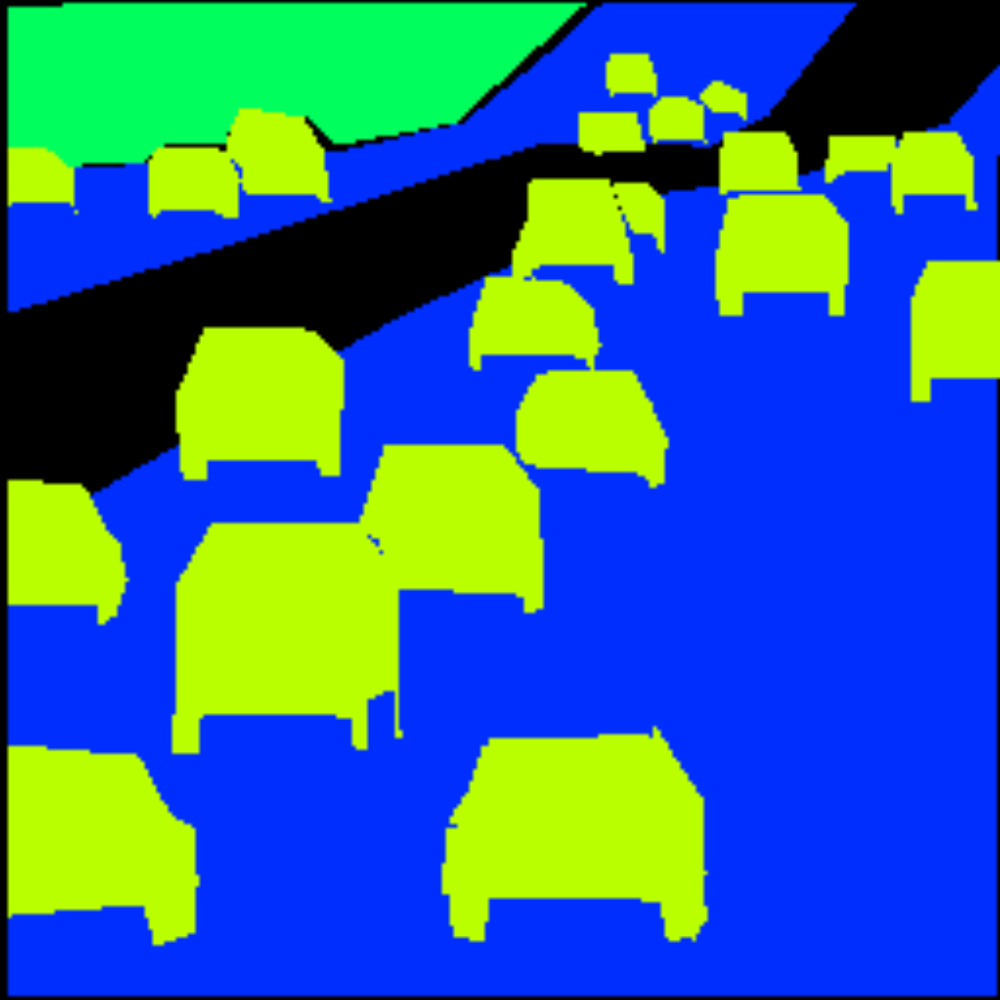}
}\hspace{2pt}
 \subfloat[]{
   \includegraphics[height=55px]{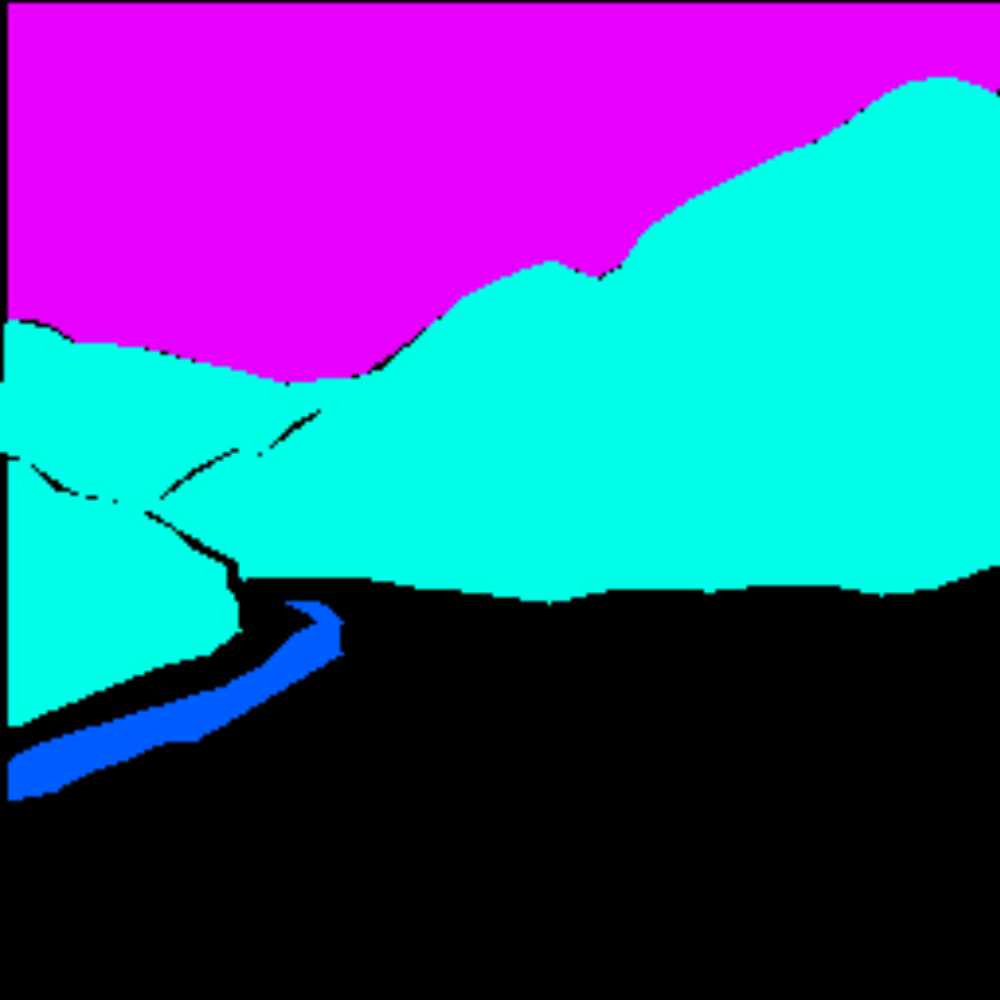}
}\hspace{16pt}
 \subfloat[]{
   \includegraphics[height=55px]{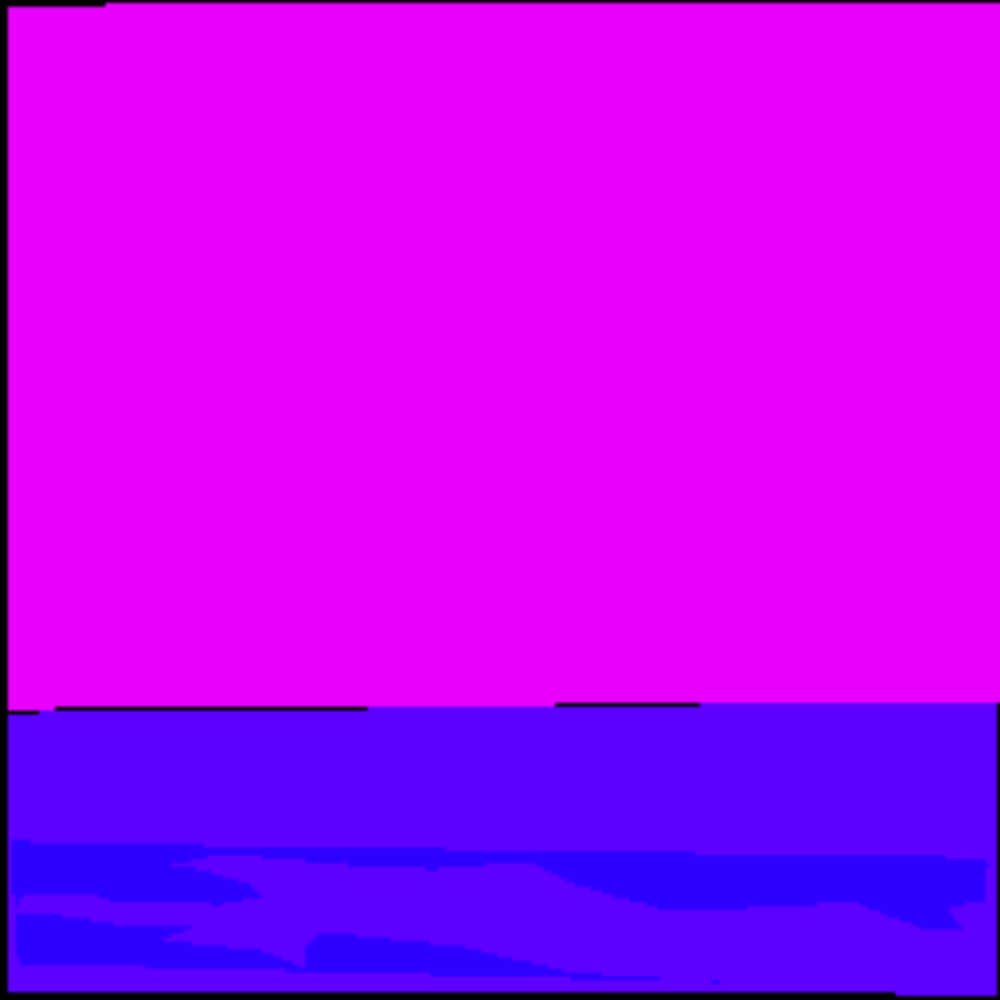}
}}
\vspace{-45pt}
\\
\subfloat[]{
   \raisebox{17pt}{\rotatebox[origin=t]{90}{$\color{white}|$SGD}}   %%% 0.75in is half of figure height 1.5in
\hspace{2pt}
 \subfloat[]{
   \includegraphics[height=55px]{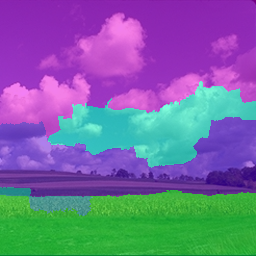}
}\hspace{2pt}
 \subfloat[]{
   \includegraphics[height=55px]{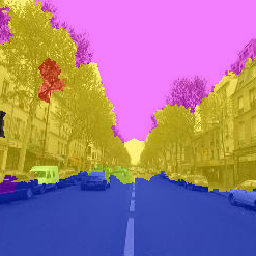}
}\hspace{2pt}
 \subfloat[]{
   \includegraphics[height=55px]{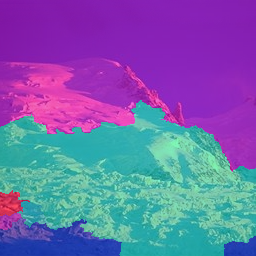}
}\hspace{2pt}
 \subfloat[]{
   \includegraphics[height=55px]{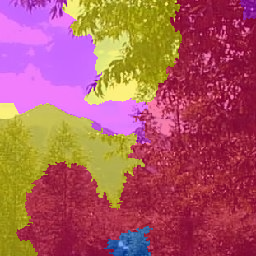}
}\hspace{2pt}
 \subfloat[]{
   \includegraphics[height=55px]{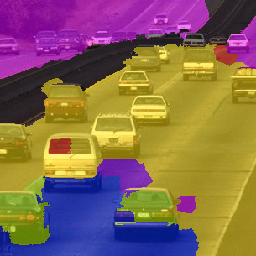}
}\hspace{2pt}
 \subfloat[]{
   \includegraphics[height=55px]{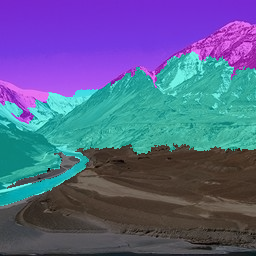}
}\hspace{16pt}
 \subfloat[]{
   \includegraphics[height=55px]{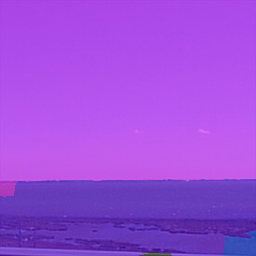}
}}
\vspace{-45pt}
\\
\subfloat[]{
   \raisebox{17pt}{\rotatebox[origin=t]{90}{$\color{white}|$int+nrl}}   %%% 0.75in is half of figure height 1.5in
\hspace{2pt}
 \subfloat[]{
   \includegraphics[height=55px]{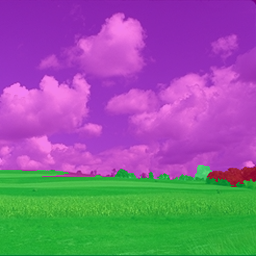}
}\hspace{2pt}
 \subfloat[]{
   \includegraphics[height=55px]{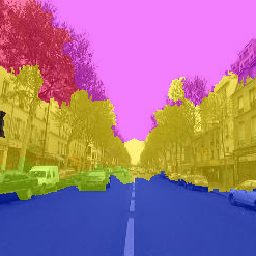}
}\hspace{2pt}
 \subfloat[]{
   \includegraphics[height=55px]{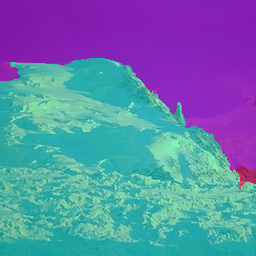}
}\hspace{2pt}
 \subfloat[]{
   \includegraphics[height=55px]{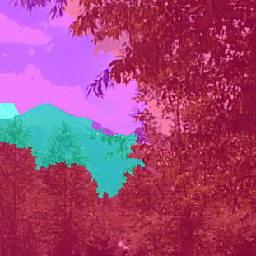}
}\hspace{2pt}
 \subfloat[]{
   \includegraphics[height=55px]{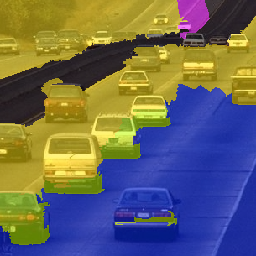}
}\hspace{2pt}
 \subfloat[]{
   \includegraphics[height=55px]{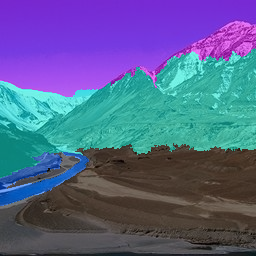}
}\hspace{16pt}
 \subfloat[]{
   \includegraphics[height=55px]{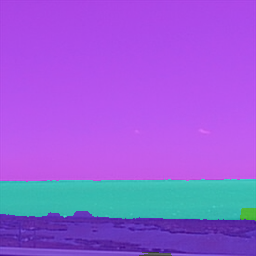}
}}
\vspace{-17pt}
\caption{Illustrative examples of the performance of SGD and int+nrl on several SIFT Flow test images. Integrated training with neural factors improves classification accuracy over subgradient descent. The last column presents a case in which our model fails to outperform SGD.}
\label{fig:results_qualitative_siftflow}
\end{figure*}

In this section, our model is analyzed on the task of image segmentation. Herein, the goal is to label different image regions with a correct class label. This is cast into a structured prediction problem by predicting all image region class labels simultaneously. There is one unary factor in underlying SSVM graphical structure for every image region, while interactions represent edges between neighboring regions. First, our model is analyzed and its different variants are compared to conventional SSVM training schemes. Second, the best performing variant is compared with state-of-the-art segmentation approaches. Our model is implemented as an extension of PyStruct \cite{JMLR:v15:mueller14a}, using Theano \cite{theano12} for GPU-accelerated neural factor optimization.

\subsection{Experimental setup}

\begin{table*}[!t]
\setlength{\tabcolsep}{3pt}
\caption{MSRC-21 class, pixel-wise, and class-mean test accuracy (in \%) for different models}\label{tab:results}
\centering
\vspace{5pt}
\begin{tabular}{l|ccccccccccccccccccccc|cc}
%\hline
& & & & & & & & & & & & & & & & & & & & & & & \\[-8pt]
				& \hspace{0.5pt} \rot{building} & \rot{grass} & \rot{tree} & \rot{cow} & \rot{sheep} & \rot{sky} & \rot{aeropl.} & \rot{water} & \rot{face} & \rot{car} & \rot{bicycle} & \rot{flower} & \rot{sign} & \rot{bird} & \rot{book} & \rot{chair} & \rot{road} & \rot{cat} & \rot{dog} & \rot{body} & \rot{boat} \hspace{0.5pt} & \rot{\bf pixel} & \rot{\bf class} \\[2pt]
\hline
& & & & & & & & & & & & & & & & & & & & & & & \\[-8pt]

unary & \hspace{0.5pt} 15 & 60 & 52 & \hphantom{0}8 & 10 & 68 & 35 & 46 & 12 & 21 & 21 & 42 & \hphantom{0}9 & \hphantom{0}2 & 36 & \hphantom{0}0 & 21 & 14 & \hphantom{0}5 & \hphantom{0}6 & \hphantom{0}1 \hspace{0.5pt} & \hspace{0.5pt} 36.3 & 23.1 \\

CP & \hspace{0.5pt} 44 & \bf 77 & 61 & 48 & 21 & 85 & 60 & 69 & 51 & 70 & 63 & 54 & 49 & 16 & 87 & 21 & 41 & 47 & \hphantom{0}6 & 16 & \bf 33 \hspace{0.5pt} & \hspace{0.5pt} 59.4 & 48.5 \\

SGD & \hspace{0.5pt} 49 & 67 & 71 & 39 & 64 & 80 & 81 & 67 & 35 & 74 & 60 & 42 & 19 & \hphantom{0}2 & 88 & \bf 51 & 53 & 38 & \hphantom{0}4 & 31 & 26 \hspace{0.5pt} & \hspace{0.5pt} 59.2 & 49.6 \\

\hline
& & & & & & & & & & & & & & & & & & & & & & & \\[-8pt]

int+lin & \hspace{0.5pt} 48 & 76 & 83 & 67 & 73 & 94 & 78 & \bf 67 & 59 & 56 & 68 & 65 & 48 & 14 & \bf 95 & 43 & 61 & 53 & \hphantom{0}6 & 45 & 32 \hspace{0.5pt} & \hspace{0.5pt} 67.4 & 58.5\\

bif+nrl & \hspace{0.5pt} 46 & 74 & 79 & 51 & 51 & 92 & \bf 83 & 64 & 76 & 64 & 67 & 50 & 53 & \hphantom{0}9 & 83 & 34 & 42 & 42 & \hphantom{0}0 & 47 & 22 \hspace{0.5pt} & \hspace{0.5pt} 62.7 & 53.7 \\

int+nrl & \hspace{0.5pt} 53 & \bf 77 & 86 & 61 & 73 & \bf 95 & \bf 83 & 60 & \bf 87 & \bf 77 & 72 & \bf 69 & \bf 77 & 27 & 85 & 29 & \bf 67 & 46 & \hphantom{0}0 & 57 & 26 \hspace{0.5pt} & \hspace{0.5pt} 70.1 & 62.3 \\

\hline
& & & & & & & & & & & & & & & & & & & & & & & \\[-8pt]

int$^\dagger$+lin \hspace{2pt} & \hspace{0.5pt} 46 & 67 & 80 & 47 & 69 & 83 & 79 & 60 & 35 & 66 & 63 & 53 & 10 & \hphantom{0}2 & 89 & 43 & 66 & \bf 62 & \hphantom{0}4 & 45 & 17 \hspace{0.5pt} & \hspace{0.5pt} 61.2 & 51.7 \\

3-layer & \hspace{0.5pt} \bf 62 & 76 & \bf 87 & \bf 68 & \bf 77 & 94 & 81 & 66 & 84 & 65 & \bf 75 & 53 & 69 & \bf 33 & 81 & \bf 51 & \bf 67 & 58 & \bf 30 & \bf 64 & 25 \hspace{0.5pt} & \hspace{0.5pt} \bf 71.6 & \bf 65.1 \\

\hline
\end{tabular}
\end{table*}

\begin{table*}[!t]
\centering
\setlength{\tabcolsep}{3pt}
\parbox{.55\linewidth}{
\caption{KITTI class, pixel-wise, and class-mean test accuracy (in \%) for different models}\label{tab:kitti_results}
\vspace{5pt}
\centering
\begin{tabular}{l|cccccccc|cc}
%\hline
& & & & & & & \\[-8pt]
				& \hspace{0.5pt} \rot{sky} & \rot{building} & \rot{road} & \rot{sidewalk} & \rot{fence} & \rot{vegetation} & \rot{pole} & \rot{car} \hspace{0.5pt} & \rot{\bf pixel} & \rot{\bf class} \\[2pt]
\hline
& & & & & & & & \\[-8pt]

unary & \hspace{0.5pt} 75 & 63 & 59 & 29 & \hphantom{0}8 & 71 & 0 & 38 \hspace{0.5pt} & \hspace{0.5pt} 53.8 & 42.8\\

CP & \hspace{0.5pt} 84 & 76 & 75 & 11 & \hphantom{0}5 & 75 & 0 & 48 \hspace{0.5pt} & \hspace{0.5pt} 61.5 & 46.7  \\

SGD & \hspace{0.5pt} 77 & 68 & 86 & 19 & \hphantom{0}4 & 80 & 0 & 71 \hspace{0.5pt} & \hspace{0.5pt} 65.5 & 50.6  \\

\hline
& & & & & & & & \\[-8pt]

int+lin & \hspace{0.5pt} 86 & 76 & 82 & 42 & 23 & 81 & \bf 6 & 67 \hspace{0.5pt} & \hspace{0.5pt} 70.2 & 57.8\\

bif+nrl & \hspace{0.5pt} 86 & 77 & 81 & 41 & 12 & 80 & 0 & 71 \hspace{0.5pt} & \hspace{0.5pt} 70.0 & 55.9 \\

int+nrl & \hspace{0.5pt} 86 & \bf 83 & \bf 88 & 50 & 19 & 84 & 4 & 74 \hspace{0.5pt} & \hspace{0.5pt} 75.6 & 60.9 \\

\hline
& & & & & & & & & \\[-8pt]

int$^\dagger$+lin \hspace{2pt} & \hspace{0.5pt} 81 & 76 & 85 & 22 & 12 & 82 & 0 & 70 \hspace{0.5pt} & \hspace{0.5pt} 69.2 & 53.5 \\

3-layer & \hspace{0.5pt}  \bf 90 &  82 & \bf 88 & \bf 55 & \bf 28 & \bf 87 & 1 & \bf 78 \hspace{0.5pt} & \hspace{0.5pt} \bf 77.6 & \bf 63.6\\

\hline
\end{tabular}
}
\hspace{20pt}
\parbox{.32\linewidth}{
\caption{SIFT Flow pixel-wise and class-mean test accuracy (in \%) for different models}\label{tab:results_siftflow}
\vspace{5pt}
\centering
\begin{tabular}{l|cc}
%\hline
& \\[-8pt]
{\color{white}\rot{vegetation}} & \hspace{0.5pt} \rot{\bf pixel} & \rot{\bf class} \\[2pt]
\hline
& &\\[-8pt]
unary & \hspace{0.5pt} 44.7 & 7.5 \\
CP & \hspace{0.5pt} 62.5 & 13.8 \\
SGD & \hspace{0.5pt} 65.9 & 15.3 \\
\hline
& &\\[-8pt]
int+lin & \hspace{0.5pt} 70.3 & 16.2 \\
bif+nrl & \hspace{0.5pt} 68.8 & 16.1 \\
int+nrl & \hspace{0.5pt} 71.3 & 17.0 \\
\hline
& &\\[-8pt]
int$^\dagger$+lin & \hspace{0.5pt} 70.2 & 15.6 \\
3-layer & \hspace{0.5pt} \bf 71.5 & \bf 17.2 \\
\hline
\end{tabular}
}
\end{table*}

The model analysis experiments are executed on the widely-used MSRC-21 benchmark \cite{shotton2009textonboost}, which consists of $276$ training, $59$ validation, and $256$ testing images. This benchmark is sufficiently complex with its 21 classes and noisy labels, and focuses on object delineation as well as irregular background recognition. Furthermore, the experiments are executed on the KITTI benchmark \cite{ros:2015} consisting of $100$ training and $46$ testing images, augmented with $49$ training images of Kundu et al.\ \cite{augm}. This latter benchmark consists of $11$ classes, but we drop the $3$ least frequently-occurring ones as they are insufficiently represented in the dataset. Finally, the same experiment is repeated for a larger dataset, namely the SIFT Flow benchmark \cite{liu2011sift}, consisting of $33$ classes with $2488$ training and $200$ testing images.

All image pixels are clustered into $\pm 300$ regions using the SLIC \cite{Achanta:2012:SSC:2377349.2377556} superpixel algorithm. For each region, gradient (DAISY \cite{Tola10}) and color (in HSV-space) features are densely extracted. These features are transformed two times into separate bags-of-words via minibatch $k$-means clustering (once 60 gradient and 30 color words, once 10 and 5 words). The unary input vectors form $(60 + 30)$-D concatenations of the first two bags-of-words. The model's connectivity structure links together all neighboring regions via edges. The edge/interaction input vectors are based on concatenations of the second set of bags-of-words. Both $(10 + 5)$-D input vectors of the edge's incident regions are concatenated into a $(2 \times (10+5))$-D vector. Moreover, two edge-specific features are added, namely the distance and angle between adjacent superpixel centers, leading to $(2 \times (10+5) + 2)$-D interaction feature vectors.

%Furthermore, the `mlayer' model is regularized using dropout with a drop probability $p=0.5$. The class weights of the loss function in Eq.~\eqref{eq:loss} are set to the inverse of the square root of the class frequency.

Factors are trained with (regular) momentum, using a learning rate curve $\frac{\mu}{t_0 + t}$, with $\mu$ and $t_0$ parameters, and $t$ the current training iteration number as used in Algorithms~\ref{algo:duplicate} and \ref{algo:duplicate2}. The regularization, learning rate, and momentum hyperparameter values are tuned using a validation set by means of a coarse- and fine-grained grid search over the parameter spaces, yielding separate settings for the unary and pairwise factors. The linear parameters $w$ are initialized to $0$, while the neural factor parameters $\theta$ and $\gamma$ are initialized according to \cite{glorot2010understanding}, except for the top layer weights which are set to $0$. The class weights $\eta(y_i^n)$ in Eq.~\eqref{eq:loss} are set to correct for class imbalance. The model is trained using CPU-parallelized loss-augmented prediction, while the neural factors are trained using GPU parallelism.

The following models are compared: unary-only (unary), $N$-slack cutting plane training (CP) with delayed constraint generation, subgradient descent (SGD)\footnote{SGD uses bifurcated training with linear interactions, hence it could be named bif+lin.}, integrated training with neural unary and linear interaction factors (int+lin), bifurcated training with neural interaction factors (bif+nrl), and integrated training with neural unary and neural interaction factors (int+nrl).

%\footnote{Trained with similar source code to unary factor training; using the library `liblinear' gave similar results.}

Multiclass logistic regression is used as unary classifier, trained with gradient descent by cross-entropy optimization. All unary neural factors contain a single hidden layer with 256 $\tanh$-units, for direct comparison of integrated learning with upfront logistic regression training. The interaction neural factors contain a single hidden layer of 512 $\tanh$-units to elucidate the benefit of nonlinear factors, without overly increasing the model's capacity. The experiment is set up to highlight the benefit of integrated learning by restricting the unary factors to features insufficiently discriminative on their own. This deliberately leads to noisy unary classification, forcing the model to rely on contextual relations for accurate prediction. The interaction factors encode information about their incident region feature vectors to allow neural factors to extract meaningful patterns from gradient/color combinations. We deliberately encoded less information in the interaction features, such that the model cannot solely rely on interaction factors for accurate and coherent predictions.

\subsection{Results and discussion}
\label{sec:results_disc}

Accuracy results on the MSRC-21 \cite{shotton2009textonboost} test images are presented in Table~\ref{tab:results}, while Figure~\ref{fig:results_qualitative} shows a handful of illustrative examples that compare segmentations attained by SGD with int+nrl. The results of the same experiment for the KITTI benchmark \cite{ros:2015}, augmented with additional training images Kundu et al.\ \cite{augm}, are shown in Table~\ref{tab:kitti_results} and Figure~\ref{fig:kitti_results}. Qualitative results on the SIFT Flow \cite{liu2011sift} dataset are shown in Figure~\ref{fig:results_qualitative_siftflow}, while accuracy results are shown in Table~\ref{tab:results_siftflow}.

\begin{figure*}[!t]
\centering
 \captionsetup[subfigure]{labelformat=empty}
%\subfloat[]{
%   \includegraphics[scale=1]{fig/bars_sgd.eps}
%}\\[-20pt]
 \subfloat[]{
   \includegraphics[width=.85\linewidth]{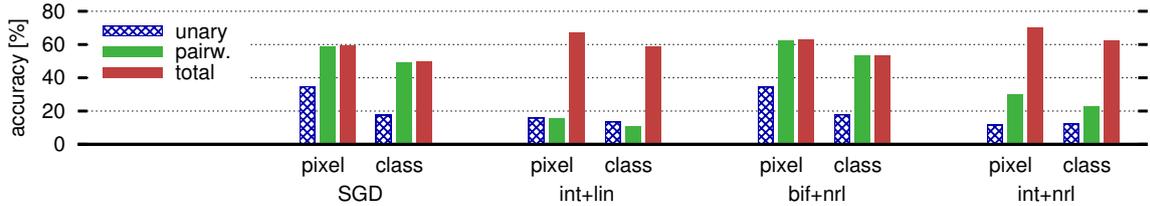}
}
\vspace{-10pt}
\caption{Visualization of the synergy between unary and interaction factors. In bifurcated training the interactions make unary factors redundant as these cannot be adapt to errors made by the interactions. In integrated training, combining both factor types leads to a higher accuracy as they can mutually adapt to each other's weight updates.}
\label{fig:factors}
\end{figure*}

\begin{table*}[!t]
\setlength{\tabcolsep}{2.5pt}
\caption{State-of-the-art comparison: MSRC-21 per-class, class-mean, and global pixel-wise test accuracy (in \%) for different models}\label{tab:sota_msrc}
\centering
\vspace{5pt}
\begin{tabular}{l|ccccccccccccccccccccc|cc}
%\hline
& & & & & & & & & & & & & & & & & & & & & & & \\[-8pt]
				& \hspace{0.5pt} \rot{building} & \rot{grass} & \rot{tree} & \rot{cow} & \rot{sheep} & \rot{sky} & \rot{aeropl.} & \rot{water} & \rot{face} & \rot{car} & \rot{bicycle} & \rot{flower} & \rot{sign} & \rot{bird} & \rot{book} & \rot{chair} & \rot{road} & \rot{cat} & \rot{dog} & \rot{body} & \rot{boat} \hspace{0.5pt} & \rot{\bf pixel} & \rot{\bf class} \\[2pt]
\hline
& & & & & & & & & & & & & & & & & & & & & & & \\[-8pt]

neural factors & \hspace{0.5pt} \bf 76 & 94 & \bf 94 & 92 & 97 & 92 & 94 & 85 & 93 & \bf 88 & 94 & 95 & 70 & \bf 78 & \bf 97 & 87 & 88 & 91 & \bf 78 & \bf 88 & \bf 63 \hspace{0.5pt} & \hspace{0.5pt} \bf 88.9 & \bf 87.4 \\

Liu et al.\ \cite{Liu:2015:CLC:2796563.2796622} & \hspace{0.5pt} 71 & 95 &  92 & 87 & \bf 98 & \bf 97 & \bf 97 & \bf 89 & \bf 95 & 85 & \bf 96 & 94 & 75 & 76 & 89 & 84 & 88 & \bf 97 & 77 &  87 &  52 \hspace{0.5pt} & \hspace{0.5pt}  88.5 & 86.7 \\

\hline
& & & & & & & & & & & & & & & & & & & & & & & \\[-8pt]

Yao et al.\ \cite{yao2012describing} & \hspace{0.5pt} 71 & \bf 98 & 90 & 79 & 86 & 93 & 88 & 86 & 90 & 84 & 94 & \bf 98 & \bf 76 & 53 & \bf 97 & 71 & \bf 89 & 83 & 55 & 68 & 17 \hspace{0.5pt} & \hspace{0.5pt} 86.2 & 79.3 \\

Lucchi et al.\ \cite{lucchi2013learning} & \hspace{0.5pt} 67 & 89 & 85 & 93 & 79 & 93 & 84 & 75 & 79 & 87 & 89 & 92 & 71 & 46 & 96 & 79 & 86 & 76 & 64 & 77 & 50 \hspace{0.5pt} & \hspace{0.5pt} 83.7 & 78.9 \\

Munoz et al.\ \cite{munoz2010stacked} & \hspace{0.5pt} 63 & 93 & 88 & 84 & 65 & 89 & 69 & 78 & 74 & 81 & 84 & 80 & 51 & 55 & 84 & 80 & 69 & 47 & 59 & 71 & 24 \hspace{0.5pt} & \hspace{0.5pt} 78 & 71 \\

Gonfause et al.\ \cite{gonfaus2010harmony} & \hspace{0.5pt} 60 & 78 & 77 & 91 & 68 & 88 & 87 & 76 & 73 & 77 & 93 & 97 & 73 & 57 & 95 & 81 & 76 & 81 & 46 & 56 & 46 \hspace{0.5pt} & \hspace{0.5pt} 77 & 75 \\

Shotton et al.\ \cite{shotton2008semantic} & \hspace{0.5pt} 49 & 88 & 79 & \bf 97 & 97 & 78 & 82 & 54 & 87 & 74 & 72 & 74 & 36 & 24 & 93 & 51 & 78 & 75 & 35 & 66 & 18 \hspace{0.5pt} & \hspace{0.5pt} 72 & 67 \\

Lucchi et al.\ \cite{lucchi2012structured} & \hspace{0.5pt} 41 & 77 & 79 & 87 & 91 & 86 & 92 & 65 & 86 & 65 & 89 & 61 & 76 & 48 & 77 & \bf 91 & 77 & 82 & 32 & 48 & 39 \hspace{0.5pt} & \hspace{0.5pt} 73 & 70 \\

\hline
\end{tabular}
\end{table*}

The results show that unary-only prediction is very inaccurate (pixel-wise/class-mean accuracy of 36.3/23.1\% for the MSRC-21 dataset, 53.8/42.8\% for the KITTI dataset, and 44.7/7.5\% for the SIFT Flow dataset). The reason for this is that unary features are not sufficiently distinctive to allow for differentiation between classes due to their low dimensionality. Accurate predictions are only possible by taking into account contextual output relations, demonstrated by the increased accuracy of CP (MSRC-21: 59.4/48.5\%; KITTI: 61.5/46.7\%; SIFT Flow: 62.5/13.8\%) as well as SGD (MSRC-21: 59.2/49.6\%; KITTI: 65.5/50.6\%; SIFT Flow: 65.9/15.3\%). These structured predictors learn linear relations between image regions, which allows them to correct errors originating from the underlying unary classifier. However, the unary factor's linear weights $w$ have only limited capability for error correction in the opposite direction, due to the fact that the SSVM cannot alter the unary classifier parameters post-hoc.

%Note that although the class-mean accuracy of int+lin is lower than CP, the model still performs better as the both models are trained for pixel-wise accuracy.

Using an integrated training approach such as int+lin, in which the SSVM is trained end-to-end, improves accuracy (MSRC-21: 67.4/58.5\%; KITTI: 70.2/57.8\%; SIFT Flow: 70.2/15.6\%) over the bifurcated procedures CP and SGD. Although neither the unary or interaction features are very distinctive, the integrated procedure updates parameters in such a way that both factor types have a unique discriminative focus. Their synergistic relationship ultimately results in higher accuracy. To better compare SGD (which uses 8, 21, and 33 logistic regression outputs as unary input features for the different benchmarks) with int+lin, we also depict the accuracy (MSRC-21: 61.2/51.7\%; KITTI: 69.2/53.5\%; SIFT Flow: 70.2/15.6\%) of a model (int$^\dagger$+lin) with only 8, 21, and 33 unary hidden units for the KITTI, MSRC-21, and SIFT Flow dataset, rather than 256 units. The 2.0/2.1\% (MSRC-21), 3.7/2.9\% (KITTI), and 4.3/0.3\% (SIFT Flow) increases in accuracy over SGD further illustrates the benefit of integrated learning and inference over conventional bifurcated SSVM training.

Another insight gained by the results is that accuracy increases when replacing linear interaction factors of conventional SSVMs with neural factors, i.e., int+nrl (MSRC-21: 70.1/62.3\%; KITTI: 75.6/60.9\%; SIFT Flow: 71.3/17.0\%) and bif+nrl (MSRC-21: 62.7/53.7\%; KITTI: 70.0/55.9\%; SIFT Flow:68.8/16.1\%) outperform int+lin (MSRC-21: 67.4/58.5\%; KITTI: 70.2/57.8\%; SIFT Flow: 70.3/16.2\%) and SGD (MSRC-21: 59.2/49.6\%; KITTI: 65.5/50.6\%; SIFT Flow: 65.9/15.3\%) respectively. This increase can be attributed to the higher number of parameters, as well as the added nonlinearities in combination with correct regularization. The model has greater generalization power, allowing the factors to extract more complex and meaningful interaction patterns. Neural factors offer great flexibility as they can be stacked to arbitrary depths. This leads to even higher generalization, as indicated by the increased accuracy (MSRC-21: 71.6/65.1\%; KITTI: 77.6/63.6\%; SIFT Flow: 71.5/17.2\%) of the deeper 3-layer (int+nrl) model. Herein both unary and interaction factors are 3-hidden-layer neural networks consisting of 256 and 512 units (rectified linear units for MSRC-21 and KITTI and $\tanh$ units for SIFT Flow) in each layer respectively. Our model can thus easily be extended, for example by letting neural factors represent the fully-connected layer in convolutional neural networks. As such, it serves as a foundation for more complex structured models.

All methods converge within 600 epochs, with one epoch taking approximately 12.62 seconds for the MSRC-21 dataset, 4.35 seconds for the KITTI dataset, and 197.27 seconds on the SIFT Flow dataset for the int+nrl algorithm. Since the implementation of our algorithm is not optimized for speed, these values can be further reduced by better exploitation of CPU parallelism.

Figure~\ref{fig:factors} illustrates the synergy between unary and interaction factors achieved through both integrated and bifurcated training, exercised on the MSRC-21 dataset. The bars depict model test accuracy when using only unary or pairwise factors, by setting either the pairwise or unary factors respectively to a zero factor value, thus $\langle w, \varphi_{I}(x,y) \rangle$ or $\langle w, \varphi_{U}(x,y) \rangle = 0 \ \forall y \in \mathcal{Y}$. Although the unary factors alone perform well in bifurcated training, nearly all accuracy can be attributed to the interactions. A possible explanation is that both types essentially learn the same information. The interactions correct errors of the underlying classifier and ultimately make unary factors redundant. In integrated training, neither the unary or interaction factors alone attain a high accuracy, but the combination of both does.

We explain this synergistic relationship with an example: Unary factors assign to a region of class \textit{A}, a second-to-highest factor value to class \textit{A}, a highest value to class \textit{B}, and a low value to class \textit{C}. The interactions also assign a second-to-highest value to class \textit{A}, but a highest value to class \textit{C}, and a low value to class \textit{B}. Independently both factors incorrectly predict the region of class \textit{A} as belonging to class \textit{B} or class \textit{C}. However, when combined they correctly assign a highest value to class \textit{A}. In the figure, bifurcated training only shows limited signs of factor synergy, as the optimization procedure is insufficiently able to steer unary and pairwise parameters in different directions, which causes them have a similar discriminative focus. This observation leads us to believe that integrated learning and inference results in higher accuracy by synergistic unary/interaction factor optimization. Both factor types are no longer optimized for independent accuracy, but mutually adapt to each other's parameter updates, which results in enhanced predictive power.

In addition to the previous experiments, the viability of our neural factor model is shown through comparison with the closely related work of Liu et al.\ \cite{Liu:2015:CLC:2796563.2796622} on the MSRC-21 dataset. Liu et al.\ make use of features extracted from square regions of varying size around each superpixel, through means of a pretrained convolutional neural network. We compare our model with theirs by using overfeat features \cite{sermanet2013overfeat} in a similar fashion, trained on individual regions. Furthermore, the model settings have been altered with respect to the previous experiments. More specifically, 1,000 SLIC superpixels are utilized for the over-segmentation preprocessing step, enforcing superpixel connectivity and merging any superpixel with a surface area below a particular threshold. DAISY gradient and HSV color features are extracted according to a regular lattice, and clustered via minibatch $k$-means clustering. Next, the same type of features are extracted for each individual pixel, leading to unary and pairwise factor feature vectors. Moreover, the $(x,y)$-position of the superpixel (median-based) center is included in the unary feature vectors, while the distance and angle between the two superpixel centers is encoded into the interaction feature vectors. The neural factors are represented by multilayer neural networks using $\tanh$-units, trained according to our Algorithm~\ref{algo:duplicate2}, using conventional momentum and single image-sized batches per gradient update. Classes are balanced by weighing them with the inverse of the class frequency. The results are presented in Table~\ref{tab:sota_msrc}, which indicate that our model is capable of performing on par with the current state-of-practice, when used in conjunction with more advanced methods, e.g., overfeat features. Moreover, similar to Liu et al.\ \cite{Liu:2015:CLC:2796563.2796622}, we have compared our model with other less closely related methods for completeness, for which the results are shown below the horizontal line in Table~\ref{tab:sota_msrc}.

\section{Conclusion}

A structured prediction model that integrates back-propagation and loss-augmented inference into subgradient descent training of structural support vector machines (SSVMs) is proposed. This model departs from the traditional bifurcated approach in which a unary classifier is trained independently from the structured predictor. Furthermore, the SSVM factors are extended to neural factors, which allows both unary and interaction factors to be highly nonlinear functions of input features. Results on a complex image segmentation task show that end-to-end SSVM training, and/or using neural factors, leads to more accurate predictions than conventional subgradient descent and $N$-slack cutting plane training. Results show that our model serves as a foundation for more advanced structured models, e.g., by using latent variables, learned feature representations, or complexer connectivity structures.

\section*{Acknowledgments}
Rein Houthooft is supported by a Ph.D. Fellowship of the Research Foundation - Flanders (FWO). Many thanks to Brecht Hanssens and Cedric De Boom for their insightful comments.

%\section*{References}

\bibliography{biblio.bib}

\end{document}